%% file: main.tex
\documentclass[10pt,twocolumn,letterpaper]{article}

\usepackage{cvpr}              % To produce the CAMERA-READY version

\definecolor{cvprblue}{rgb}{0.21,0.49,0.74}
\usepackage{soul}
\usepackage{tcolorbox}
\usepackage[pagebackref=true,breaklinks=true,colorlinks,bookmarks=false]{hyperref}
\hypersetup{linkcolor=[rgb]{0.7,0.1,0.1}}
\hypersetup{citecolor=cvprblue}

\usepackage{tikzducks}
\usepackage{bm}
\usepackage{multirow}
\usepackage{soul}
\usepackage{wrapfig}
\usepackage{microtype}

\usepackage{caption}
\usepackage{subcaption}
\usetikzlibrary{calc,arrows,arrows.meta,backgrounds,tikzmark,shapes.geometric, decorations.pathreplacing,decorations.markings, spy,matrix,positioning}
\usepackage[ruled]{algorithm2e} % For algorithms

\SetAlFnt{\small}
\SetAlCapFnt{\small}
\SetAlCapNameFnt{\small}
\SetAlCapHSkip{0pt}
\usepackage{pifont}
\usepackage{colortbl}
\usepackage{tcolorbox}
\renewcommand{\captionlabelfont}{\footnotesize}
\newlength\savewidth\newcommand\shline{\noalign{\global\savewidth\arrayrulewidth
  \global\arrayrulewidth 1pt}\hline\noalign{\global\arrayrulewidth\savewidth}}

\def\paperID{7413} % *** Enter the Paper ID here
\def\confName{CVPR}
\def\confYear{2025}

\title{ChatHuman: Chatting about 3D Humans with Tools
}

\newcommand{\methodname}{\textcolor{black}{\mbox{ChatHuman}}\xspace}
\newcommand{\model}{\textcolor{black}{\mbox{ChatHuman}}\xspace}

\newcommand*{\new}[1]{\textcolor{black}{#1}}

\input{config/commands}

\input{config/authors}
\begin{document}

\twocolumn[{
\renewcommand\twocolumn[1][]{#1}
\maketitle
\begin{center}
    \captionsetup{type=figure}
    \vspace{-7mm}
    \includegraphics[width=\textwidth]{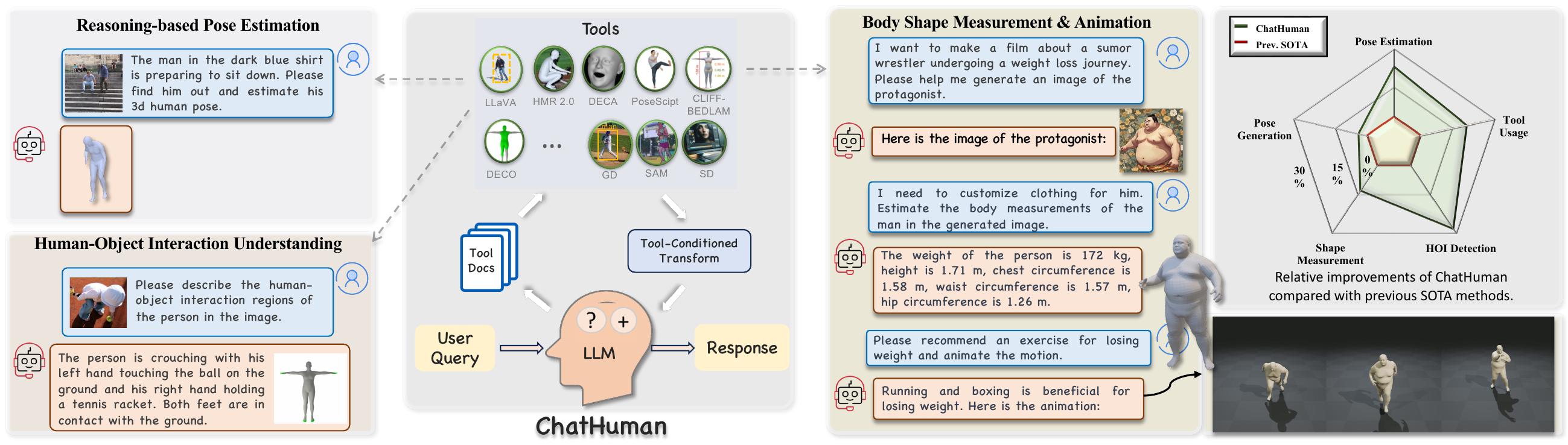}
    \vspace{-3mm}
    \captionof{figure}{\model is a LLM-based agent that uses a multimodal LLM to exploit and combine tools, discriminate their results, and integrate the results to solve tasks related to 3D Humans. 
    }
    % \vspace{-1mm}
    \label{fig:teaser}
\end{center}
}]

\input{sections/0_abstract}

\input{sections/1_introduction}

\input{sections/2_related}

\input{sections/method}
\input{sections/4_experiment}

\input{sections/5_conclusion}
\input{sections/6_acknowledgement}

{
    \small
    \bibliographystyle{ieeenat_fullname}
    \bibliography{main}
}
\input{sections/X_supp}


\end{document}

%% file: config/commands.tex
\newcommand{\qheading}[1]{\noindent\textbf{#1:}}

\usepackage{lipsum}
\usepackage{balance}

\definecolor{citecolor}{HTML}{0071bc}
\definecolor{frontcolor}{HTML}{325ea5}
\definecolor{sidecolor}{HTML}{a58b77}
\definecolor{DeltaColor}{rgb}{0.039,0.73,0.71}
\definecolor{SigmaColor}{rgb}{0.98,0.45,0.0}
\definecolor{AlphaColor}{rgb}{0,0,0.8}
\definecolor{BetaColor}{rgb}{0.8,0,0.8}
\definecolor{GammaColor}{rgb}{0.514,0.34,0.224}
\definecolor{EpsilonColor}{rgb}{0.353,0.725,0.906}
\definecolor{PurpleColor}{HTML}{bca5ea}
\definecolor{OrangeColor}{rgb}{0.914,0.541,0.0.141}
\definecolor{GreenColor}{rgb}{0.137,0.573,0.565}
\definecolor{RedColor}{rgb}{0.949,0.275, 0.224}
\definecolor{LightCyan}{rgb}{0.88,1,1}
\definecolor{Gray}{gray}{0.91}

\newcommand{\numtools}{26\xspace}

\newcommand{\moveToSupMat}[1]{\begin{comment}#1\end{commment}}
\newcommand{\supmat}{\textcolor{magenta}{{Sup.~Mat.}}\xspace}

\newcommand{\colorRef}[1]{\textcolor{red}{#1}} %
\usepackage[capitalize]{cleveref}
\crefname{figure}{\colorRef{Fig.}}{\colorRef{Figs.}}
\Crefname{figure}{\colorRef{Figure}}{\colorRef{Figures}}
\crefname{section}{\colorRef{Sec.}}{\colorRef{Secs.}}
\Crefname{section}{\colorRef{Section}}{\colorRef{Sections}}
\Crefname{table}{\colorRef{Table}}{\colorRef{Tables}}
\crefname{table}{\colorRef{Tab.}}{\colorRef{Tabs.}}
\Crefname{equation}{\colorRef{Equation}}{\colorRef{Equations}}
\crefname{equation}{\colorRef{Eq.}}{\colorRef{Eqs.}}

%% file: config/authors.tex
\author{Jing Lin\textsuperscript{3,4,$\star$} \quad %
Yao Feng\textsuperscript{2,3,$\star$} \quad %
Weiyang Liu\textsuperscript{1,5} \quad
Michael J.~Black\textsuperscript{1} 
\\
\textsuperscript{1}Max Planck Institute for Intelligent Systems, T\"ubingen~~~\textsuperscript{2}Stanford University
\\\textsuperscript{3}Meshcapade~~~\textsuperscript{4}Tsinghua University~~~\textsuperscript{5}University of Cambridge
\\
\href{https://chathuman.github.io/}{\texttt{\small chathuman.github.io}}
}

%% file: sections/0_abstract.tex
\begin{abstract}
% \vspace{-4mm}
% new story
% Problem:  
\def\thefootnote{}\footnotetext{\textsuperscript{$\star$}Equal contribution. This work was done while YF and JL were at Meshcapade.}
Numerous methods have been proposed to detect, estimate, and analyze properties of people in images, including 3D pose, shape, contact, human-object interaction, and emotion. While widely applicable in vision and other areas, such methods require expert knowledge to select, use, and interpret the results. 
To address this, we introduce ChatHuman, a language-driven system that integrates the capabilities of specialized methods into a unified framework.  ChatHuman functions as an assistant proficient in utilizing, analyzing, and interacting with tools specific to 3D human tasks, adeptly discussing and resolving related challenges. Built on a Large Language Model (LLM) framework, ChatHuman is trained to autonomously select, apply, and interpret a diverse set of tools in response to user inputs. 
% Adapting LLMs to 3D human tasks presents challenges, including the need for domain-specific knowledge and the ability to interpret complex 3D outputs. 
Our approach overcomes significant hurdles in adapting LLMs to 3D human tasks, including the need for domain-specific knowledge and the ability to interpret complex 3D outputs. 
The innovations of ChatHuman include leveraging academic publications to instruct the LLM on tool usage, employing a retrieval-augmented generation model to create in-context learning examples for managing new tools, and effectively discriminating between and integrating tool results by transforming specialized 3D outputs into comprehensible formats. 
Experiments demonstrate that ChatHuman surpasses existing models in both tool selection accuracy and overall performance across various 3D human tasks, and it supports interactive chatting with users. ChatHuman represents a significant step toward consolidating diverse analytical methods into a unified, robust system for 3D human tasks. Code and data are available at ~\tt\href{https://chathuman.github.io}{chathuman.github.io}.

\end{abstract}

%% file: sections/1_introduction.tex
% \vspace{-5}
\vspace{-0.4cm}
\section{Introduction}
Research on 3D humans has progressed rapidly, resulting in the creation of many tools that can perform tasks like estimating a human's 3D pose from a single image~\cite{hmr,hmr2,cliff,spin, lin2023osx}, predicting face/body shapes~\cite{shapy,DECA:Siggraph2021}, capturing emotions~\cite{DECA:Siggraph2021,danvevcek2022emoca}, and identifying regions of touch/contact~\cite{deco,muller2021self}, generating human poses from text descriptions~\cite{posescript}, and animating human images~\cite{zhu2024champ}.  
Each of these tools, however, focuses on a specific problem, functioning as isolated ``specialists''. 
Moreover, these separate tools cannot benefit from the expertise of others, and combining them to solve more complex tasks requires significant domain expertise. 
Ideally, we would have a single model that can adaptively leverage different tools to solve complex 3D human-related problems while offering intuitive user interaction through natural language input.
Recent work such as ChatPose \cite{feng2024chatpose} has taken initial steps in this direction, unifying pose generation, estimation, and general understanding within an LLM framework. 
Unfortunately, ChatPose lacks the accuracy of the best specialist methods.

To address these issues, we build a multi-modal LLM, called \methodname, that specializes in using digital human modeling tools, enabling it to autonomously interpret instructions and complete diverse tasks related to 3D humans; see Fig.~\ref{fig:teaser}.
Specifically, we teach an LLM to use a wide range of specialized human-related models for tasks like 3D pose estimation, emotion recognition, contact reasoning, and more, effectively extending the LLM’s capabilities to the domain of 3D humans. 
This goes beyond providing a natural-language interface to these tools, as the LLM can use its broad understanding of humans to augment tool results or to analyze and integrate their outputs, providing better responses than any single tool alone.

\iffalse
To address complex, real-world, tasks, we would like a ``generalist'' model that can solve a wide range of problems with equal or better accuracy than the specialists. 
Such a system should be able to use a wide range of specialist tools, know how to apply them to appropriate tasks, and be able to synthesize the results of several tools to solve new problems. For example, reasoning-based pose estimation (Fig.~\ref{fig:teaser}) can be addressed through the combination of text-guided detection (LLaVA \cite{llava}), cropping, and human pose estimation tools (HMR2 \cite{hmr2}), rather than relying on a single tool alone.  
Recent work such as ChatPose \cite{feng2024chatpose} has taken initial steps in this direction, unifying pose generation, estimation, and general understanding within an LLM framework. However, this approach is limited to trained tasks and tools, lacking the flexibility to adapt to new tools or tasks as they emerge.
\fi 

% Our solution 
With \methodname, we introduce a novel approach by finetuning an LLM to act as an agent that autonomously calls appropriate tools based on user inputs, completing tasks and enhancing responses with tool-generated results. Similar in spirit, recent works have employed off-the-shelf or fine-tuned LLMs for tasks like basic vision (e.g., Visual ChatGPT~\cite{wu2023visual-chatGPT}), mobile applications (e.g., AppAgent~\cite{yang2023appagent}), biology (e.g., AmadeusGPT \cite{ye2023amadeusgpt}) and system automation (e.g., GPT4Tools~\cite{gpt4tools}). 
Our work, however, differs by focusing specifically on the unique challenges of 3D human understanding. This domain requires precise, specialized terminology and a nuanced understanding of 3D-specific tools, which conventional LLMs lack. 
To teach the network this specialized terminology, we do what we would do as humans -- we have the LLM read the papers describing the methods.
%~\yao{Unfortunately (or fortunately?), now ChatGPT can read papers..and it even knows ChatHuman} 
Even with that knowledge, the LLM needs to understand the task goals, select an appropriate tool or tools, interpret results, and resolve differing results. These skills are all beyond the abilities of general LLMs.
%, complex task resolution requires both an understanding of task goals and expertise in tool selection and interpretation, particularly when interpreting tool outputs is beyond the current capabilities of general LLMs.

% Solution
To address these challenges, we design the following training pipeline: 1) We utilize relevant literature about the tools to familiarize the LLM with domain knowledge, helping it know when and how to use these tools; 
2) After using a tool, the LLM evaluates the reliability of the outcome using its ``judgment" and compares different methods to identify the most reliable results; 
3) It combines these results with its general knowledge to create response.  
This pipeline represents several key innovations, laying a foundation for LLMs to effectively handle complex, tool-driven 3D human tasks. 

\textbf{Retrieval-Augmented Tool Use}: 
Details about tools are typically present in corresponding academic paper. We give the LLM access to these papers and demonstrate that ``reading the paper'' improves tool use performance. We further analyze which paper sections are most effective for instructing tool use. 
When encountering a new tool, users often turn to the user guide for assistance. 
We compile the complete documentation for these tools and utilize a paper-based Retrieval-Augmented Generation (RAG) mechanism to improve the LLM's understanding and management of new tools. This means that, although the LLM has not encountered such tools during fine-tuning, it can still effectively use the tools with the help of the paper-based RAG mechanism.  
% \yao{new about tool graph here, make it very weak now}
In some cases, tasks require combining multiple tools. To address a broader range of tool usage scenarios, we employ a graph-based invocation system, which includes a node for single-tool use, a chain for sequential tool execution, and a DAG for multi-tool combinations as shown in Fig.~\ref{fig:tool_combination}.

\textbf{3D Human-Related Tool Result Integration}: 
Analyzing outputs from tools is crucial, as these outputs, such as body meshes, model parameters (e.g., SMPL pose), or motion sequences, are highly varied and complex. To make these results compatible with our LLM analysis system, we convert them into visual formats that the LLM can easily interpret. Guided by Cognitive Load Theory \cite{Sweller2011CognitiveLT}, we present these outputs as multiple-choice options, streamlining the selection process and enhancing the LLM’s effectiveness in handling 3D human-related tasks. Combined with the LLM’s extensive general knowledge, these integrated results enable it to generate sophisticated responses about 3D humans.

Specifically, \model consists of a multimodal LLM LLaVA~\cite{llava}, and \numtools tools involving 3D Humans and general vision tasks.  
The LLM is finetuned to use these tools and incorporate their results.  
User requests can be in the form of text descriptions, images or other 3D information (if applicable), and the model produces text descriptions, images, or other 3D outputs after tool reasoning. 
Extensive evaluations demonstrate that \model not only surpasses previous models in tool-use accuracy but also improves performance on various human-related tasks. 
It achieves this by reasoning about multiple outputs, evaluating their veracity, and combining them with its own knowledge.

Summarizing, our key contributions include:
(1) a framework that leverages LLMs to interact with users and address human-centric tasks using specialist tools; 
(2) a scientific-paper-based RAG mechanism that ensures precise tool use by comprehending tool descriptions from scholarly articles, enhancing tool applications and interactions; and 
(3) the integration of tool outcomes with LLMs, enabling the LLM to effectively explain tool results and interact with users. Additionally, the LLM is fine-tuned to distinguish between optimal and suboptimal tool results, improving overall accuracy. 
% (4) A dataset about 3D humans for LLMs.. ~\yao{can we claim dataset as contribution?}
\methodname achieves superior performance in tool use and human-related tasks compared with other LLM-based methods or task-specific methods. 
The code, trained models, and datasets are available for research purposes.

%% file: sections/2_related.tex
\vspace{-2mm}
\section{Related work}
\qheading{3D Humans}
%There is an extensive literature related to research on 3D humans, which we only sample here. 
Many 3D human analysis tasks  leverage parametric models like SMPL~\cite{smpl}, SMPL-X~\cite{smplx}, or GHUM~\cite{xu2020ghum} for the body, BFM~\cite{bfm09} or FLAME~\cite{FLAME:SiggraphAsia2017} for faces, and MANO~\cite{MANO:SIGGRAPHASIA:2017} for hands. These models enable the representation of the human body, face, and hands as low-dimensional vectors, facilitating subsequent applications in estimation and generation. 
Estimation of human pose and shape either relies on optimization methods~\cite{smplify,eft}  or regression methods \cite{hmr,hmr2,zhang2021pymaf,cliff,hybrik,spin,shapy,lin2023osx,pixie,rong2021frankmocap}, which estimate body shape and pose parameters from a single image. 
Similarly, face reconstruction methods \cite{DECA:Siggraph2021,deng2019accurate,tewari2017mofa} 
estimate shape and expression parameters of the face model from single images. 
The analysis of contact, vital for understanding human-environment interaction and social touch, has seen recent attention~\cite{muller2021self,deco,han2023chorus}. 
Generative modeling techniques such as PoseScript~\cite{posescript} and PoseFix~\cite{delmas2023posefix} provide methods for synthesizing and correcting 3D human poses based on textual descriptions, while language-to-3D generation methods~\cite{cao2023dreamavatar,hong2022avatarclip,zhang2023dreamface} facilitate 3D avatar creation. Additionally, numerous methods generate human motions \cite{petrovich23tmr,jiang2023drop,li2023object,tevet2023human,tevet2022motionclip,aberman2019learning,posegpt,li2021learning}, and recent language-to-video models are even able to generate humans moving \cite{blattmann2023stable,zhu2024champ}.
% For human behavior understanding, prior work focuses on classifying action labels in video sequences \cite{rajasegaran2023benefits,pan2021actor} or recognizing human emotions \cite{danvevcek2022emoca,usman2017using}.

These basic methods excel in their respective scenarios, but are typically treated in isolation.
When mature, such tools are often incorporated into software systems for animators that require significant domain knowledge. 
In contrast, recent generative models provide language interfaces to image, video, and 3D generation tools, making them accessible to novices.
Until recently, such language-based control has not been possible for 3D humans. 
ChatPose~\cite{feng2024chatpose} makes a step in this direction, unifying pose generation, estimation, and an LLM's general understanding into one model, but remains limited in its task capabilities.
%in full-body pose tasks. 
In contrast, our model integrates the performance of \numtools 3D human-related tasks into a single, LLM-based model. %
\model enables non-experts to solve real-world tasks by invoking appropriate tools and adding an extra layer of language-driven understanding that effectively leverages the tool outputs.

\qheading{Large Language Models and Tool Use} To expand LLM capabilities without expensive retraining, recent work has focused on enabling them to use specialized tools. 
In this approach, a tool library is constructed and LLMs act as planners to coordinate tool usage.
Various tools have been adopted, \eg, vision modules~\cite{wu2023visual-chatGPT,surismenon2023vipergpt,yang2023mm}, mobile applications~\cite{yang2023appagent}, community tools~\cite{shen2023hugginggpt}, special tools~\cite{ye2024amadeusgpt}, and system tools~\cite{gpt4tools}. 
However, general-purpose LLMs often lack a deep understanding of specific tools, especially those requiring domain knowledge. 
To address this, recent work \cite{wang2024mllmtool,TPTU-v2,AnyTool} proposes to fintune general-purpose LLMs (\eg, LLaMA~\cite{touvron2023llama}, LLaVA~\cite{liu2023visual,liu2023llavaplus}) with domain-specific tool-use data. Some methods use additional tool documentation to improve accuracy \cite{hsieh2023tool,yuan2024easytool}, while others compose different tools to accomplish complex tasks \cite{wu2024toolplanner,ruan2023tptu,kong2023tptu} .
Distinct from previous work, \model focuses on 3D human tasks through language interaction by leveraging off-the-shelf human-related tools.

\qheading{Retrieval Augmented Generation} RAG \cite{lewis2020retrieval,gao2023retrieval,zhao2023retrieving} is a technique to enhance generative tasks by retrieving relevant information from external databases, allowing for continual knowledge update. Here, we design a RAG mechanism to facilitate the use of new unseen tools.

%% file: sections/method.tex
\section{Method}
% Finally, in Section~\ref{sec:data_construction}, we discuss the construction of a large-scale training dataset with instruction-following data for tool usage and feedback.
% This dataset is designed to train the model to effectively use tools and integrate their results to generate accurate responses. 

% \begin{figure}[t]
%   \centering  
%   \includegraphics[width=\linewidth]{figures/pipeline_v2.pdf}
%   \vspace{-6.5mm}
%   \caption{\footnotesize Method Overview. Given a user query, the multimodal LLM-based agent adopts a paper-based RAG mechanism to determine whether to employ tools and identify the optimal way to utilize them. After applying the tools, the tool results are transformed into a text or visual format and fed back to the agent to formulate responses. \yao{update figure with tool combination with nodes/tree}}
%   \vspace{-2.5mm}
%   \label{fig:method}
% \end{figure}

%\vspace{-1mm}
\subsection{Overall Pipeline} 
\label{sec:training} 
% As depicted in Figure~\ref{fig:method}, 
\methodname consists of a multimodal LLM $f_{\phi}(\cdot)$, along with a set of 3D human-related functions ${f_{\theta_1}(\cdot), f_{\theta_2}(\cdot), ...}$. These functions serve as tools for various tasks, such as 3D human pose estimation, pose generation, and 3D face reconstruction. 
Our model takes input text queries $X_q$, images $X_v$, and optionally $X_m$ representing other 3D human-related modalities (e.g., SMPL parameters for 3D human poses).
Then it invokes tools and integrates their results to generate outputs as text $Y_t$, images $Y_v$, or 3D human-related modalities $Y_m$.  
% The pipeline consists of two steps: automatically invoking tools based on user queries and integrating tool results to enhance the LLM’s output.  
% Given a user query, the multimodal LLM-based agent adopts a paper-based RAG mechanism to determine whether to
% employ tools and identify the optimal way to utilize them. After applying
% the tools, the tool results are transformed into a text or visual format and
% fed back to the agent to formulate responses
\begin{figure}[t]
  \centering  
  \includegraphics[width=\linewidth]{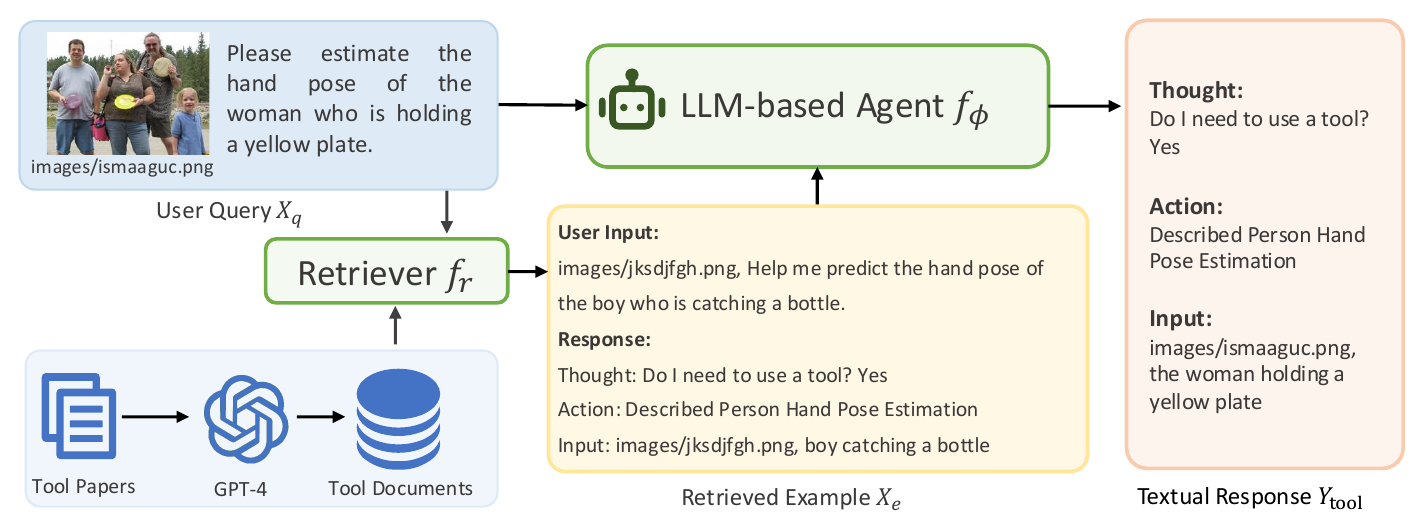}
  \vspace{-6mm}
  \caption{Paper-based Retrieval-Augmented Tool Use. We feed academic papers describing each tool to GPT-4 to build a document for each tool. 
  During inference, given a user query, a relevant sample is retrieved from the documents and provided to the LLM-based agent as an in-context example to improve tool use accuracy.}
  \vspace{-2.5mm}
  \label{fig:rag_tool_usage}
\end{figure}

\subsection{Retrieval-Augmented Tool Usage} 
\label{sec: RAG} 
Teaching LLMs to decide when and how to use tools effectively is a significant challenge. A basic approach \cite{wu2023visual-chatGPT, gpt4tools} might involve including tool usage scenarios and input arguments within the LLM prompt, represented as $Y_\text{tool} = f_{\phi}(X_q, X_t)$, where $X_t$ denotes tool definitions. However, this approach often falls short for specialized tools, especially given the variety of advanced tools for 3D human tasks. Many tools require background knowledge for correct use and have multiple application scenarios. For instance, the HMR tool \cite{hmr2} may be queried with requests like, ``Can you estimate this person’s pose?”, ``What are the SMPL parameters?”, or ``Provide the 3D mesh of this person.” Capturing all possible usage scenarios succinctly in a prompt is difficult, and as tools proliferate, prompt descriptions become unwieldy.
% It is challenging to capture all possible usage scenarios and necessary background knowledge succinctly in a prompt. As the number of tools grows, prompt descriptions become lengthy and complex, complicating tool selection and combination, especially for tools the LLM was not trained on. 
To address these challenges, we introduce paper-based Retrieval-Augmented Generation (RAG)~\cite{RAG} and build a tool graph for tool combination. 
As shown in Fig.~\ref{fig:rag_tool_usage}, we feed academic papers associated with each tool into GPT-4, prompting it to summarize the tool's functions and generate possible user queries for tool activation. These papers, with their rich background and detailed instructions, enable the generation of user queries that cover diverse use cases. By combining these queries with each tool's structured arguments, we compile a document of question-answer pairs for each tool’s operation. Figure \ref{fig:rag_tool_usage} provides an example from one of these documents. 
These documents serve as an auxiliary knowledge base during inference, from which we retrieve a relevant example $X_e$ in response to a user query $X_q$. The retrieval process matches the text embedding of the query with embeddings in the tool documents using a text embedding model \cite{su2022one}. The retrieved sample is then presented to the agent $f_\phi$ as an in-context learning example:
\begin{equation} 
X_e = f_r(X_q),~~{Y}_\text{tool} = f_\phi(X_q, X_e, X_t),
\end{equation} 
where $f_r$ is the retrieval function, and $Y_\text{tool}$ is a textual description of the tool invocation, specifying tool selection, names, and input arguments for tool calls. 

% The model then invokes the tools and obtains the tool results $Y_m$. 

\noindent\textbf{Graph-based Tool Invocation.} 
\label{sec:tool_combination} 
Note that the tool use description $Y_{tool}$ varies depending on task settings, as shown for a single tool case in Fig.~\ref{fig:rag_tool_usage}. However, some complex tasks require combining multiple tools.
To handle this, we introduce a graph-based mechanism for tool invocation. We construct a tool graph with three structure types: nodes (single tool calls), chains (tool sequences for dependent tasks), and directed acyclic graphs (DAGs) \cite{shen2023taskbench} for complex multi-branch operations. For each user query, the model predicts an appropriate tool graph and invokes the tools accordingly. Examples of tool graphs are shown in Fig.~\ref{fig:tool_combination}. 
 
\begin{figure}[t]
  \centering  
  \includegraphics[width=\linewidth]{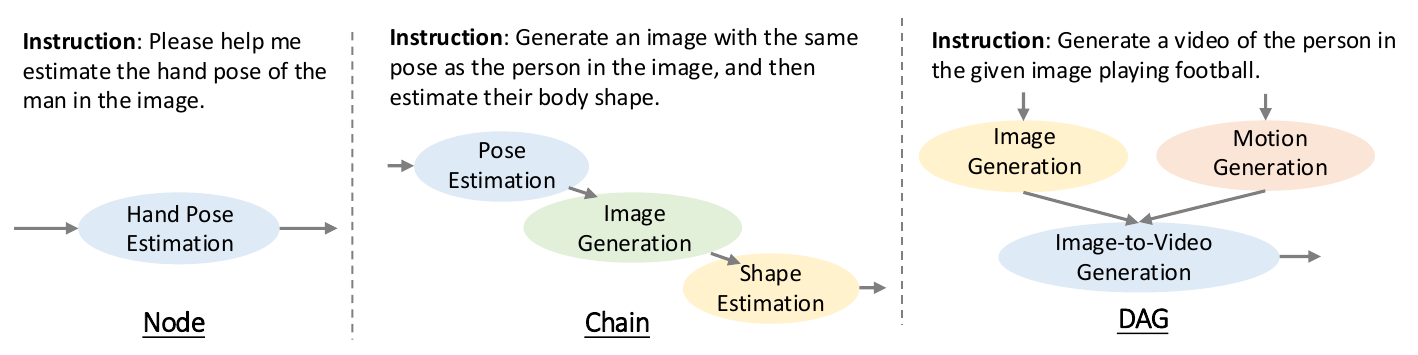}
  \vspace{-6.5mm}
  \caption{Examples of tool use graphs. Tool use patterns include: a Node, which uses a single tool; a Chain, which requires sequential tool execution; and a DAG, which combines multiple tools. }
  \vspace{-3mm}
  \label{fig:tool_combination}
\end{figure}

\begin{figure}[t]
  \centering  
  \includegraphics[width=\linewidth]{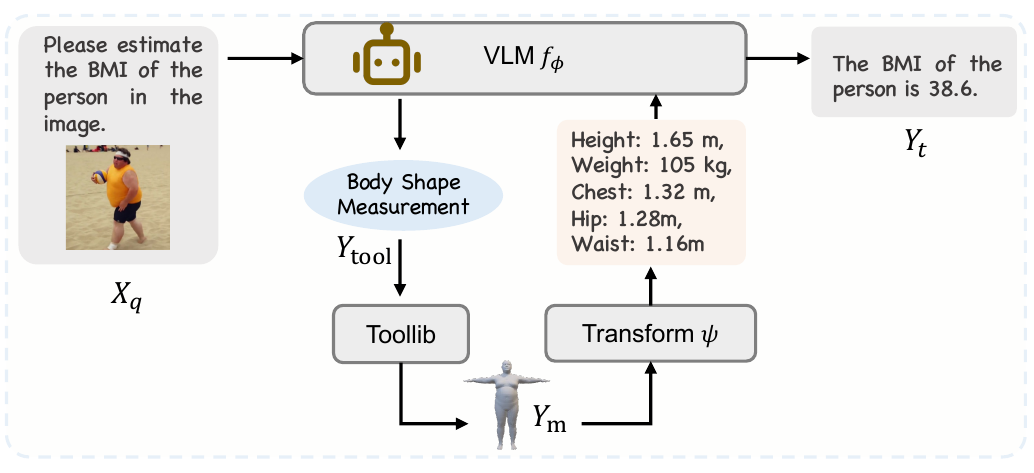}
  \vspace{-6.5mm}
  \caption{Illustration of tool result integration. After tool invocation, results are transformed into VLM-compatible representations. The VLM then incorporates these transformed outputs to generate accurate, informative responses to user queries.}
  \vspace{-2.5mm}
  \label{fig:tool_integration}
\end{figure}

\subsection{Tool Result Integration} 
\label{sec: tool feedback} 
After using tools, integrating their results is essential to effectively engage with users and solve problems. However, outputs from different tools vary widely, appearing as language, images, or vectors (like SMPL poses), which can challenge current multimodal LLMs, such as LLaVA~\cite{llava}, that process only text and images. To utilize these varied results and enhance the LLM’s understanding of 3D humans, thereby improving its ability to apply world knowledge to problem-solving, we introduce a tool-conditioned transformation, $\Psi(\cdot)$. As shown in Fig.~\ref{fig:tool_integration}, this transformation converts tool outputs $Y_m$ into textual or visual formats the LLM can process. 
For example, we transform the vertex-wise contact label from DECO \cite{deco} into body part-level descriptions using SMPL's \cite{smpl} vertex-to-part mapping dictionary, and render the mesh generated by PoseScipt \cite{posescript} into an RGB image using rendering techniques. See \supmat for more details.  
The transformed results are then merged with the user query as context for response generation: 
\begin{equation}
Y_t=f_\phi(X_q, \Psi(Y_m)).
\end{equation} 
In scenarios where multiple tools can address a request (Fig.~\ref{fig:tool_discrimination}), we present outcomes as multiple-choice questions, prompting the model to select the most relevant answer:
\begin{equation}
Y_t = f_\phi(X_q, \Psi(Y_{m1}),\Psi(Y_{m2}),...),
\end{equation} 
where $Y_{mi}$ denotes the $i$-th tool result. 
Since different tools have different failure modes, this process enables ChatHuman to identify the best method case by case, producing more accurate output than any individual method alone.

\begin{figure}[t]
  \centering  
  \includegraphics[width=\linewidth]{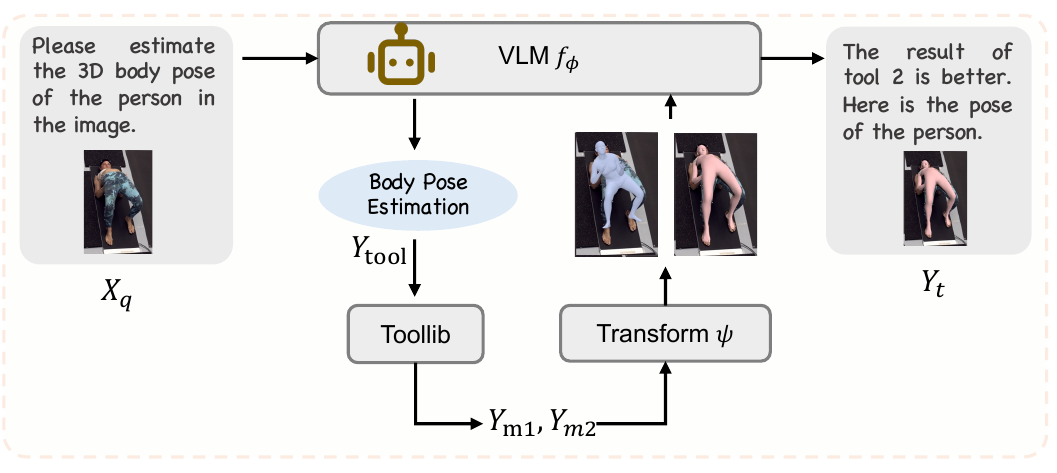}
  \vspace{-6.5mm}
  \caption{Illustration of tool results discrimination. When multiple plausible tools exist for a task, ChatHuman discriminates and chooses the best result as the final response.}
  \vspace{-2.5mm}
  \label{fig:tool_discrimination}
\end{figure}

\begin{figure*}[h]
  \centering 
  \vspace{-2mm}
  \includegraphics[width=.99\textwidth]{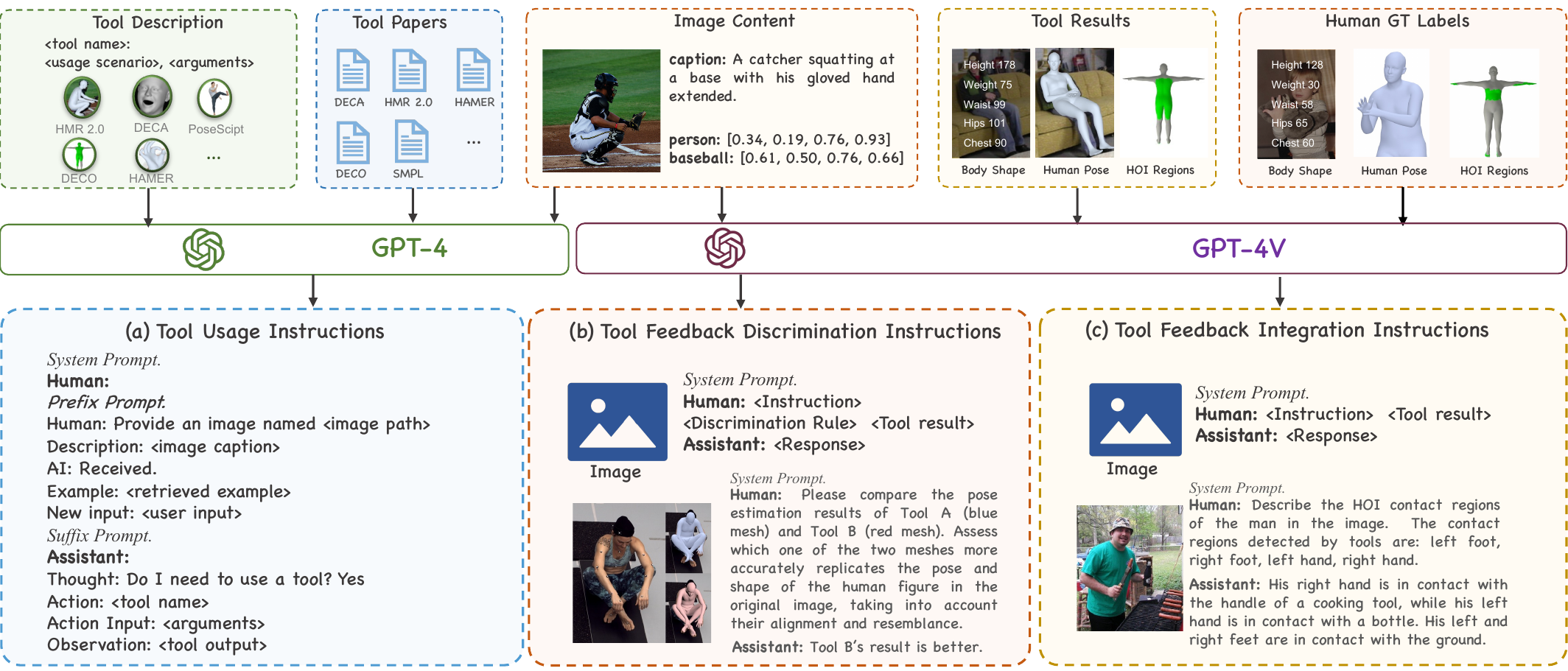}
  \vspace{-1.5mm}
  \caption{ Illustration of our instruction-following data construction pipeline. We construct tool usage and feedback data by providing GPT-4 with various tool-related information, image content, and ground truth labels. \textcolor{gray}{Gray} text shows some example instructions.}
  \vspace{-2mm}
  \label{fig:data_construction}
\end{figure*}

% \begin{figure*}[t]
%   \centering  
%   \includegraphics[width=\textwidth]{figures/pose_generation_description.pdf}
%   \vspace{-7mm}
%   \caption{\footnotesize Examples of the instruction-following data for discriminating pose generation and pose description results.}
%   \vspace{-2mm}
%   \label{fig:pose_gen_descrimination}
% \end{figure*} 

\new{
\subsection{Training Data Construction}
\label{sec: training_data}
\textbf{Tool Usage Instruction-following Data.} To teach the LLM-based agent to correctly use tools, we construct 90K instruction-response pairs about tool usage. Following GPT4Tools \cite{gpt4tools}, we provide GPT-4 \cite{gpt4} with a textual description of COCO training images \cite{coco} and a tool-related prompt containing a tool description.  To improve efficiency, we first prompt GPT-4 to summarize paper content, re-articulate tool functions, and enumerate 50 potential user queries for tool activation (see Fig.~\ref{fig:data_construction}(a))}.

\noindent\new{\textbf{Tool Feedback Instruction-following Data.} To help the multimodal LLM model discriminate and integrate the tool results, we construct 88K pairs of instruction-following data based on existing datasets 3DPW \cite{threedpw}, MOYO \cite{moyo}, PoseScript \cite{posescript} and SHAPY \cite{shapy} (see  Fig.~\ref{fig:data_construction}(b)(c) ).
}
Please see \supmat for more details about data construction.

\subsection{Model Training}
% To finetune our network, we construct instruction-following data about tool use and result integration by prompting GPT4 with task specific datasets \cite{coco, posescript, threedpw,moyo}.  
Once we have data, we use LoRA~\cite{hu2021lora} to finetune the LLM $f_{\phi}(\cdot)$ with the cross entropy loss. More specifically, with the ground truth tool invocation labels $\hat{Y}_{tool}$ and response label $\hat{Y}_{t}$, we optimize the model using the following objective function:
 $\mathcal{L} = \mathbf{CE}(\hat{Y}_\text{tool}, Y_\text{tool}) + \mathbf{CE}(\hat{Y}_t, Y_t)$, where $\mathbf{CE}$ denotes the cross-entropy loss. See \supmat for details.

\vspace{-1mm}

%% file: sections/4_experiment.tex
\section{Experiments and Results}
\subsection{Implementation Details} 
We use LLaVA-1.5 \cite{llava} as the VLM backbone, with CLIP \cite{clip} for vision encoding and Vicuna \cite{vicuna} for the LLM backbone. For retrieval, we adopt INSTRUCTOR \cite{su2022one} for text embedding and utilize Chroma's vector similarity searching algorithm to identify relevant examples. 
To preserve the generalization of the pretrained multi-modal LLM, we use LoRA \cite{hu2021lora} to perform efficient finetuning, with rank 128 and alpha  256. %Alternatively, orthogonal finetuning~\cite{liu2023parameter,qiu2023controlling} can be used to improve performance. 
\new{We implement tool utilization with LangChain~\cite{langchain}, which enables automatic parsing of tool names and input parameters, followed by tool invocation}
Optimization uses AdamW \cite{adamw}, with a learning rate of 2e-4 and weight decay of 0. All models are finetuned over 2 epochs with a mixture of tool usage, tool feedback, and LLaVA multimodal instruction-tuning data, using 8 Nvidia A100-80G GPUs with the DeepSpeed \cite{rasley2020deepspeed} engine. 
Unless otherwise specified, we use LLaVA-1.5-7B as the base model for the ablation study. 
We support \numtools tools, as listed in Tab.~\ref{tab:tool_name}.
\input{tables/10_tool_list}
% Details of the training data are in \supmat.

% \vspace{-1.75mm}
% \subsection{Datasets} 
% \vspace{-.4mm}
% \noindent\textbf{Tool Usage Instruction-Tuning Data.} To teach the agent to correctly use tools, we construct 90K instruction-response pairs about tool usage. Our tool library consists of \numtools human-related tools. We further construct a validation and test set for evaluation. The validation set has 1000 samples with the same tools as the training set, while the test set includes 689 samples related to 3 tools not presented during training. To ensure the difference between the training and test sets, we use varied image caption sources for input prompts. More details are provided in the \supmat. 

% \vspace{1mm}
% \noindent\textbf{Tool Feedback Instruction-Tuning Data.} To help the multimodal LLM model discriminate and integrate the tool results, we construct 88K instruction-following data based on existing 3D human datasets, including 61K tool result discrimination instructions built with MoYo \cite{moyo}, 3DPW \cite{threedpw}, and PoseScript \cite{posescript}, and 27K tool result integration instructions from SHAPY \cite{shapy} and DECO \cite{deco}. 

\input{tables/01_tool_usage_sota}
\vspace{-1.75mm}
\subsection{Evaluation on Tool Usage} 
\vspace{-.5mm} 
\textbf{Tool Usage Benchmark.} To evaluate tool usage accuracy, we construct a validation and test set. The validation set has 1000 samples with the same tools as the training set, while the test set includes 689 samples related to 3 tools unseen during training. Split of seen and unseen tools are detailed in Table~\ref{tab:tool_name}.
% The unseen tools include Remove Someone From the Photo, Replace Someone from the Photo, and Targeted Hand Pose Estimation, as shown 
 Similar to our training data construction, we feed a textual description of COCO validation set image, a tool description, and some examples summarized from the tool paper into GPT-4 and prompt it to generate instruction-following data. We use the image description captioned by LLaVA \cite{llava} instead of the original image captions to ensure differences between training and test sets. Finally, we manually check all question-answering pairs for accuracy.

\noindent\textbf{Baselines.} We compare our method with Visual ChatGPT \cite{wu2023visual-chatGPT} and GPT4Tools \cite{yang2023gpt4tools} on the proposed evaluation set and report 5 metrics proposed in GPT4Tools \cite{yang2023gpt4tools}. See \supmat for details of the metrics. 
For Visual ChatGPT, we experiment with two versions of GPT: ``gpt-3.5-turbo-1106'' and ``gpt-4-turbo-preview''. For GPT4Tools, we adopt the official pretrained 13B model. For a fair comparison, we also finetune GPT4Tools with our training data using the official training code and obtain a variant, GPT4Tools-FT. 

Table \ref{tab:tool_usage} shows that the original GPT4Tools struggles on our benchmark due to mismatches between its tools and our human-centric ones. Visual ChatGPT-4 exhibits impressive tool use accuracy, showing its powerful zero-shot ability to follow a standardized format and use tools accurately. \methodname consistently outperforms other methods, particularly with tools not seen in training, thanks to our paper-based RAG mechanism. 
% \yao{Supmat: details of the unseen tools, visualize tool combination of specific task}

\vspace{-0.15cm}

\subsection{Evaluation on 3D Human Related Tasks}

% In the following, we evaluate the performance of \methodname on  representative human-related tasks. 
\begin{figure}[t]
  \centering  
  %\vspace{-1.75mm}
  \includegraphics[width=0.98\linewidth]{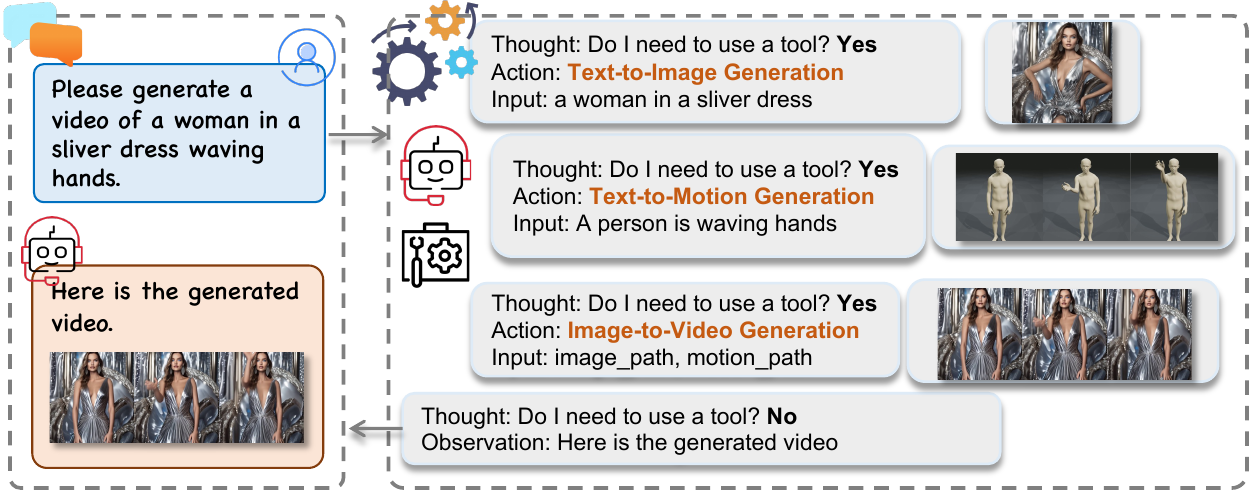}
  \vspace{-3mm}
  \caption{Visualization of Animation Processing. Left: Conversation between the user and ChatHuman. Right: ChatHuman automatically calls tools to solve the task. ChatHuman can handle tasks beyond the capabilities of individual tools.} 
  \vspace{-6mm}
  \label{fig:animation_visualization}
\end{figure}

\noindent\textbf{Character Animation.}
 \model employs tools for text-to-mot\-ion and image-to-video generation. We demonstrate how these tools are utilized to interact with users and reason about motions based on conversations in Fig.~\ref{fig:animation_visualization} and Fig.~\ref{fig:teaser}. 
\model can also tackle tasks that cannot be resolved with a single tool. For instance, text-to-human video generation poses significant challenges due to the complexity of motion. Therefore, another option is to first generate a motion sequence via text-to-motion generation, then apply a video generation model conditioned on this sequence. The internal processing within \model, detailing how it analyzes and solves tasks, is visualized in Fig.~\ref{fig:animation_visualization}. We also compare our text-to-video generation results with those of Pika\footnote{We use the demo available at \url{https://pika.art/} (as of May 2025) to obtain the results.}. The qualitative comparisons are shown in Fig.~\ref{fig:rpe_comparison}.

\vspace{.75mm}
\noindent\textbf{Pose Estimation.} Following ChatPose \cite{feng2024chatpose}, we evaluate the performance of our method on both classical and reasoning-based pose estimation (RPE) tasks. MPJPE, PA-MPJPE, and MPJRE on the 3DPW \cite{threedpw} and RPE \cite{feng2024chatpose} benchmarks are reported. For the reasoning-based pose estimation task, \methodname first grounds a human based on a textual description and feeds it into the pose estimation tool to get the result.  As shown in Table \ref{tab:exp_hmr}, \methodname achieves comparable performance to the task-specific models on the classical pose estimation task. For reasoning-based human pose estimation, which involves both reasoning ability and advanced human pose estimation ability, \methodname outperforms both task-specific and multi-modal LLM methods by a large margin ($34.6\% \downarrow$ in MPVPE). As shown in Fig.~\ref{fig:rpe_comparison}, only our method achieves a satisfactory result. The multimodal LLM competitor ChatPose finds the correct person but fails to obtain an accurate pose due to its limited perception ability, while the task-specific tool does not match the correct person due to the lack of reasoning ability. This demonstrates the advantage of \methodname, which combines task-specific tool use expertise with the general reasoning ability of an LLM. 

\input{tables/02_hmr_sota}

\begin{figure}[t]
  \centering  
  \vspace{-1.75mm}
  \includegraphics[width=\linewidth]{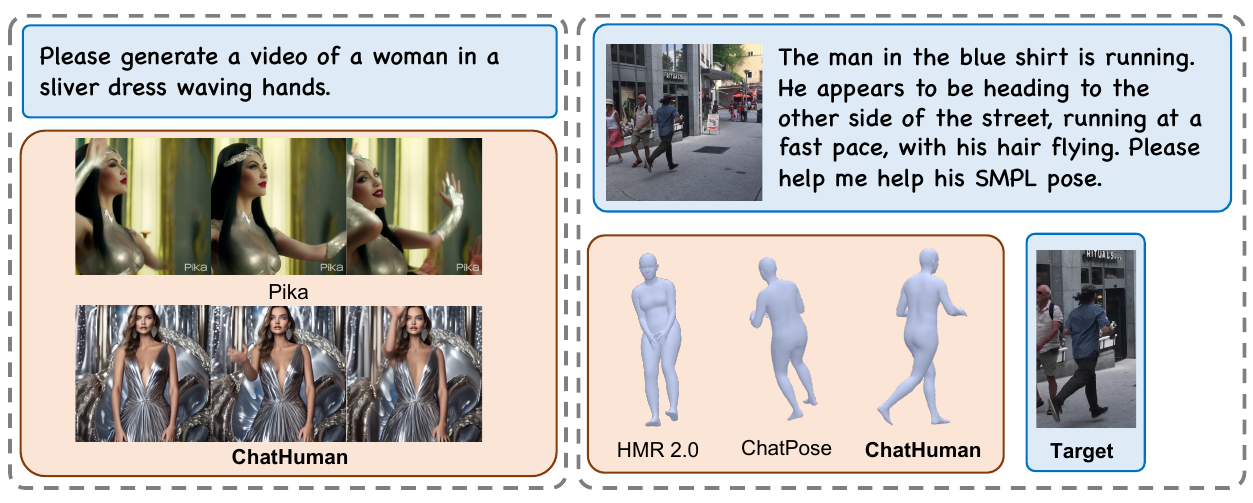}
  \vspace{-7mm}
  \caption{Left: Comparison to Pika$^1$ on text to video generation. Right: Qualitative comparison with ChatPose \cite{feng2024chatpose} and HMR 2.0 \cite{hmr2} on reasoning-based human pose estimation task.}
  \vspace{-4.5mm}
  \label{fig:rpe_comparison}
\end{figure}

\vspace{.75mm}
\noindent\textbf{Pose Generation.} Here we evaluate the pose generation capability of \methodname on the classical text-to-pose generation task and the speculative pose generation task (SPG) \cite{feng2024chatpose}. Following previous work \cite{posescript, feng2024chatpose}, we report the text-to-pose recall rate $R^{T2P}$ and pose-to-text recall rate $R^{P2T}$ of the retrieval models trained on real poses and evaluated on generated poses. For the SPG task, \methodname first rephrases the indirect pose descriptions into explicit ones and adopts PoseScript (journal version) \cite{posescript} to generate a pose.
As shown in Table \ref{tab:exp_pose_generation}, our method archives comparable performance to the SOTA methods on both benchmarks. In contrast, the previous LLM-based method, ChatPose, performs poorly on the classical pose generation benchmark, while the task-specific model, PoseScript, lags in the SPG benchmark due to limited reasoning ability.
% \yao{Supmat: example of indirect pose description into explicit ones}

\input{tables/03_pose_generation}

\vspace{.75mm}
\noindent\textbf{Body Shape Measurement.} We evaluate the body shape measurement accuracy of \methodname. We randomly sample 100 images from the HBW validation set \cite{shapy} and compare our method with a multimodal LLM, LLaVA \cite{llava}, and a SOTA body shape estimation method, CLIFF-BEDLAM \cite{bedlam}. For LLaVA and \methodname, we ask them the same question to inquire about the height, weight, chest, waist, and hip circumferences of a person and then prompt GPT-3.5 to extract the value from the model output. The details of the question and prompt are available in \supmat 
CLIFF-BEDLAM predicts the body shape parameters, which are then converted to measurements based on the shape-to-measurement function from SHAPY \cite{shapy}. Anthropometric measurement errors are reported in 
%Table \ref{table:shape_estimation}. 
Table~\ref{table:shape_estimation_hoi}(a). \methodname achieves superior performance in most measurements, outperforming the multimodal LLM competitor LLaVA by $42\%$ and CLIFF-BEDLAM by $15.7\%$ in average metric accuracy.

\vspace{.75mm}
\noindent\textbf{Human-Object Interaction (HoI).} We evaluate the human-object interaction understanding ability of \methodname on the DECO \cite{deco} test set. The ground truth (GT) labels are obtained by converting the vertex-level contact labels into body part-level contact labels with SMPL's vertex-to-part mapping dictionary. Given a human image, we ask the multimodal LLM to detect the body parts contacting objects and prompt GPT-3.5 to extract the body part labels from the answer. Subsequently, we compare the predicted body parts with the GT label and compute the average detection precision, recall rate, and F1 Score.  
%Table \ref{table:hoi}
Table~\ref{table:shape_estimation_hoi}(b) shows that \methodname achieves SOTA precision and F1 score, demonstrating superior human-object interaction understanding ability. Notably, although LLaVA has a high recall rate, its precision and F1 score are rather poor, which means that it tends to predict all the body parts to be in contact with objects.

\input{tables/shape_HOI}

\subsection{Ablation Study}

\noindent\textbf{Paper-based RAG Mechanism.} To improve tool use accuracy, we design a paper-based RAG mechanism. We perform a breakdown ablation to investigate the effect of each component and their interactions. The baseline model, created by removing the RAG operation and trained with instruction-following data without paper content, is compared in Table \ref{table:tool_ablation}(a). The baseline model's success rate (SR) 
is 0.96 for seen tools and 0.82 for unseen tools. Adding RAG increases the SR for unseen tools to 0.89, demonstrating its effectiveness in zero-shot settings. Further incorporating  articles into training data boosts the performance: the successful rate of arguments ($\text{SR}_\text{args}$) rises from 0.93 to 0.95 for the seen tools and 0.84 to 0.94 for the unseen tools. This suggests that scholarly articles can help create high-quality instruction-following data and tool documents due to their detailed use instructions and diverse application scenarios. For a detailed analysis of each component's effect on instructing tool usage, please see \supmat

\noindent\textbf{Multiple Tools Invocation.} One of the advantages of using a VLM as an agent is its powerful generalization capacity. To test the robustness and generalization ability of ChatHuman, we conduct the following ablation study. During training, we only include the tool graphs with no more than three tools, while during evaluation, the user queries might need up to five tools to solve. Table \ref{table:tool_ablation}(b) depicts the results. As shown, ChatHuman exhibits an excellent robustness in this out-of-domain setting (more than three tools combination) with an action accuracy higher than 90\%.

\input{tables/tool_ablation}

\input{tables/07_feedback_ablation}

\input{tables/08_discrimator_ablation} 
\vspace{1mm}
\noindent\textbf{Tool Result Integration.} We first conduct an ablation to study how the tools can enhance the human understanding capacity of multimodal LLM. The model without tools is our multimodal LLM backbone, LLaVA-1.5-7B \cite{llava}, and the model with tools is our \methodname. The quantitative results are listed in Table \ref{table:ablation_tool2agent}. When equipped with tools, the HOI contact detection F1 score increases from 0.39 to 0.63 and the average body shape measurement error declines by 38\%. These results demonstrate the effectiveness of tools in enriching the LLM's comprehension of human models and behaviors.  
 Additionally, we study whether \model can utilize its world knowledge to discriminate and improve the tool performance. We design two discrimination schemes, i.e., selection and modification, and conduct an ablation study on two human-related tasks by comparing \methodname with the SOTA task-specific tools. For the selection scheme, we experiment with the pose estimation task and select two SOTA methods, HMR 2.0 \cite{hmr2} and CLIFF-SMPLify \cite{cliff,smplify}, as our tools to generate two poses of each person. We then prompt the LLM-based agent to discriminate the results and choose the better one as the final response. Different tools excel in different scenarios and, to cover more diverse human poses and camera views, we build a new benchmark MixPose by selecting 100 images with extreme camera views from the MoYo \cite{moyo} test set, 100 full-body samples and 100 severely-truncated samples from 3DPW \cite{threedpw} test set. Details of the prompt and MixPose benchmark are in \supmat As shown in Table \ref{table:ablation_agent2tool} (a), \methodname archives a lower reconstruction error on the MixPose benchmark, validating the agent's effectiveness as a discriminator to improve tool performance. For the modification scheme, we validate on the body shape measurement task. We use CLIFF-BEDLAM \cite{bedlam} as tool and prompt the agent to discriminate and modify the tool result. The result is reported in Table \ref{table:ablation_agent2tool} (b), and Fig.~\ref{fig:body_shape_comparison}. The LLM-based agent enhances tool performance by using its general world knowledge to identify and correct unreasonable tool results, such as height and weight in Fig.~\ref{fig:body_shape_comparison} (a). 
 
 \begin{figure}[t]
  \centering  
  %\vspace{-1mm}
  \includegraphics[width=0.47\textwidth]{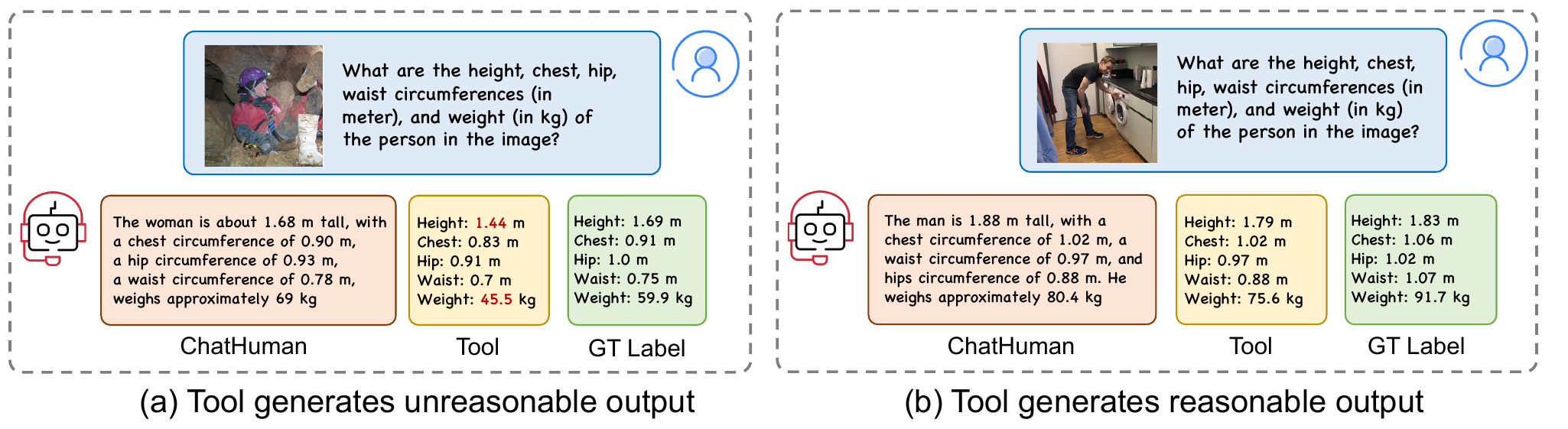}
  \vspace{-2mm}
  \caption{Illustration of how ChatHuman discriminates and integrates tool results. The Agent will fix the unreasonable tool result and integrate the reasonable tool result to generate a final response. }
  \vspace{-6mm}
  \label{fig:body_shape_comparison}
\end{figure}

% \begin{figure}[t]
%   \centering  
%   \vspace{-1mm}
%   \includegraphics[width=\linewidth]{figures/body_shape_comparison.pdf}
%   \vspace{-7mm}
%   \caption{\footnotesize Illustration about how the multimodal LLM-based agent discriminates and integrate tool results. The Agent will fix the unreasonable tool result and integrate the reasonable tool result to generate a final response. }
%   %\vspace{-.25mm}
% \label{fig:body_shape_comparison}
% \end{figure}

%% file: tables/10_tool_list.tex
\begin{table}[t]
    \vspace{-2mm}
	\centering
	\scriptsize
        \setlength{\tabcolsep}{1.75pt}
	\renewcommand{\arraystretch}{1.25}
	\renewcommand{\captionlabelfont}{\footnotesize}
	\resizebox{\columnwidth}{!}{
		\begin{tabular}{l|l|l}
			\multicolumn{1}{c|}{\textbf{Perception}} & \multicolumn{1}{c|}{\textbf{Reasoning}} & \multicolumn{1}{c}{\textbf{Generation}} \\
			\shline
             Body Pose Estimation~\cite{hmr2} & ~Selective Person Pose Detection~\cite{hmr2,llava} & ~Text-to-Pose Generation~\cite{posescript}\\
             Body Shape Measurement~\cite{bedlam} & ~Specific Person Shape Measurement~\cite{llava,bedlam} & ~Speculative Pose Generation~\cite{llava,posescript} \\
             Hand Pose Estimation~\cite{hamer} & ~\cellcolor{Gray}Targeted Hand Pose Estimation~\cite{llava,hamer} & ~Text-to-Image Generation~\cite{stable_diffusion} \\
             Face Reconstruction~\cite{DECA:Siggraph2021} & ~Described Person Face Reconstruction~\cite{llava,DECA:Siggraph2021} & ~Text-based Pose Editing~\cite{delmas2023posefix} \\
            Human Segmentation~\cite{kirillov2023segment} & ~Described Person Segmentation~\cite{llava,kirillov2023segment} & \cellcolor{Gray}{Remove Someone From The Photo}~\cite{llava,kirillov2023segment,stable_diffusion}\\
             HOI Detection~\cite{deco} & ~Selective Person Contact Estimation~\cite{llava,deco} & ~\cellcolor{Gray}Replace Someone From The Photo ~\cite{llava,kirillov2023segment,stable_diffusion}\\
             Pose Description~\cite{posescript} & ~Visual Question Answering~\cite{llava} & ~Instruct Image Using Text~\cite{stable_diffusion} \\
             Image Caption~\cite{llava} & & ~Text-to-Motion Generation~\cite{petrovich23tmr}\\
             Motion Capture~\cite{shin2023wham} & & ~Text-to-Video Generation~\cite{petrovich23tmr,stable_diffusion,zhu2024champ} \\
             & & ~Image-to-Video Generation~\cite{petrovich23tmr,zhu2024champ} \\

		\end{tabular} 
	}
    \vspace{-0.08in}
 \caption{
	%List of tools used in \model: 
    \model supports 26 human-related tools, including 9 perception tools, 10 generation tools, and 7 reasoning tools. Tools in grey are unseen tools that are not included in the training data.
}
	\label{tab:tool_name}
    \vspace{-1mm}
\end{table}

%% file: tables/01_tool_usage_sota.tex
\begin{table}[t]
	\centering
	\scriptsize
	\setlength{\tabcolsep}{1.2pt}
	\renewcommand{\arraystretch}{1.45}
	\renewcommand{\captionlabelfont}{\footnotesize}
 %\vspace{-.25mm}
	\resizebox{\columnwidth}{!}{
		\begin{tabular}{l|ccccc|ccccc}
			&  \multicolumn{5}{c|}{ Seen Tools} & \multicolumn{5}{c}{Unseen Tools}\\
			{Method} & $~~\text{SR}_\text{t}~~$ & $~~\text{SR}_\text{act}~~$ & $~~\text{SR}_\text{args}~~$ & $~~\text{SR}~~$ &~~IoU~~& $~~\text{SR}_\text{t}~~$ & $~~\text{SR}_\text{act}~~$ & $~~\text{SR}_\text{args}~~$ & $~~\text{SR}~~$ & ~~IoU~~ \\
			\shline
            GPT4Tools \cite{yang2023gpt4tools}  &  0.609 &     0.547     &     0.525   &  0.520 &    0.566 & 0.612 &     0.546   &  0.542 &    0.525 & 0.573    \\
			GPT4Tools-FT \cite{yang2023gpt4tools}  &  0.825 &     0.710     &     0.687   &  0.690 &    0.741 & 0.904 &     0.807   &  0.690 &    0.747 & 0.800    \\
			Visual ChatGPT-3.5 \cite{wu2023visual-chatGPT}  &  0.498 &     0.319     &     0.237   &  0.251 &    0.791 & 0.507 &     0.314   &  0.226 &    0.293 & 0.803    \\
           Visual ChatGPT-4 \cite{wu2023visual-chatGPT}  &  0.892 &     0.802     &     0.715   &  0.753 &    0.797 & 0.998 &     0.913   &  0.801 &    0.872 & 0.907    \\\rowcolor{Gray}
			\methodname  &  \bf 1.000 &    \bf  0.974     &    \bf  0.950   & \bf  0.970 &   \bf  0.975 & \bf 0.999 &    \bf  0.967  & \bf  0.893 &   \bf  0.954 &\bf  0.953    \\
		\end{tabular} 
	}
 \vspace{-1.1mm}
    \caption{
	            Tool use accuracy comparison. Successful rate of thought ($\text{SR}_\text{t}$), action ($\text{SR}_\text{act}$), arguments ($\text{SR}_\text{args}$), execution ($\text{SR}$), and IoU are reported. 
	}
 \vspace{-3.5mm}
	\label{tab:tool_usage}
\end{table}

%% file: tables/02_hmr_sota.tex
\begin{table}[t]
	\centering
	\scriptsize
        \setlength{\tabcolsep}{1.2pt}
	\renewcommand{\arraystretch}{1.25}
	\renewcommand{\captionlabelfont}{\footnotesize}
	\resizebox{\columnwidth}{!}{
		\begin{tabular}{l|ccc|ccc}
			&  \multicolumn{3}{c|}{ 3DPW \cite{threedpw}} & \multicolumn{3}{c}{RPE Benchmark \cite{feng2024chatpose}}\\
			{Method} & {~MPJPE $\downarrow$~} & {~PA-MPJPE $\downarrow$~} & {~MPJRE $\downarrow$~} & {~MPJPE $\downarrow$~} & {~PA-MPJPE $\downarrow$~} & {~MPJRE $\downarrow$~} \\
			\shline
            SPIN \cite{spin}  &  102.9 &     62.9     &     10.1   &  244.9 &    107.3 & 12.4  \\
            HMR 2.0 \cite{hmr2}   &     \textbf{91.0}      &    \textbf{58.4}     &  \textbf{9.2} &    225.2    &      105.1 & 12.1        \\
			\hline
                LLaVA-S \cite{llava}          &     440.8    &      205.4    & 21.8  &   490.7 & 207.4 & 21.1       \\
                LLaVA*-S \cite{llava}       &     232.1 &      101.1    &      12.8  &      - & - & -   \\
                GPT4-S \cite{gpt4}      &     322.0  &      136.7    &      16.0  &    - & - & -   \\
                LLaVA-P \cite{llava}      &     335.2    &      172.3    &  16.5 &    391.5    &      191.9 & 17.8  \\
                GPT4-P \cite{gpt4}        &     396.5    &      203.4    &      18.6   &   -  & - & -  \\
                ChatPose \cite{feng2024chatpose}     &     163.6      &     81.9    &   10.4    &      253.6     & 103.8 & 11.7     \\\rowcolor{Gray}
                \methodname     &     91.3      &     58.7    &   \textbf{9.2}    &      \textbf{147.2}     & \textbf{79.1} & \textbf{10.3}    \\
		\end{tabular} 
	}
 %\vspace{0.75mm}
 \vspace{-0.1in}
    \caption{
	            Comparison of vanilla human pose estimation and reasoning-based pose estimation on 3DPW \cite{threedpw} and RPE \cite{feng2024chatpose}.
				LLaVA* is fine-tuned with human keypoint data. ``S'' uses multimodal LLMs for keypoint detection and SMPLify~\cite{smplify} for pose optimization. ``P'' utilizes multimodal LLMs for textual pose descriptions, processed by PoseScript~\cite{posescript} to generate poses.
				MPJPE (in mm), MP-MPJPE (in mm), and MPJRE ($\times100$) are reported.
	}
	\label{tab:exp_hmr}
 \vspace{-1mm}
\end{table}

%% file: tables/03_pose_generation.tex
\begin{table}[t]
	\centering
	\scriptsize
        \setlength{\tabcolsep}{1.25pt}
	\renewcommand{\arraystretch}{1.25}
	\renewcommand{\captionlabelfont}{\footnotesize}
	\resizebox{\columnwidth}{!}{
		\begin{tabular}{l|ccc|ccc|ccc|ccc}
			& \multicolumn{6}{c|}{ PoseScript \cite{posescript}} & \multicolumn{6}{c}{SPG Benchmark \cite{feng2024chatpose}}\\
			{Method} & \multicolumn{3}{c}{$R^{P2T}$ $\uparrow$} & \multicolumn{3}{c|}{$R^{T2P}$ $\uparrow$} & \multicolumn{3}{c}{$R^{P2T}$ $\uparrow$} & \multicolumn{3}{c}{$R^{T2P}$ $\uparrow$} \\
			\shline
            PoseScript \cite{posescript}~~     &    ~40.4~ & ~52.3~ & ~65.0~  &  ~41.4~ & ~\textbf{54.1}~ & ~65.9~   &    ~1.5 ~& ~3.5~ & ~6.2~  &    ~ 1.4~ & ~2.3~ & ~5.1~      \\
                ChatPose \cite{feng2024chatpose}       &    17.6  & 25.3 & 35.8     & 28.0 & 39.0& 54.4  &      \textbf{3.3} & \textbf{5.5} & {8.2} &      ~\textbf{3.5} & {5.8} & \textbf{11.0} \\
				LLaVA-P \cite{llava}       &    -  & - & -     & - & -& - &      2.1 & 4.0 & 7.1 &      ~2.1 & 3.3 & 6.1 \\
				GPT4-P \cite{gpt4}       &    -  & - & -     & - & -& - &     2.7&4.7&9.2 &    2.7  & 5.3 & 8.2  \\\rowcolor{Gray}
                \methodname      &    \textbf{41.8} & \textbf{52.6} & \textbf{65.1}     & \textbf{42.1} & 52.3 & \textbf{66.5}
				&      3.2 & {5.0} & \textbf{9.9} & \textbf{3.5} & \textbf{6.5} & 10.6 \\
		\end{tabular} 
	}
        \vspace{-2.5mm}
	\caption{
	            Comparison of classical and speculative pose generation on PoseScript \cite{posescript} and SPG \cite{feng2024chatpose}. ``P” denotes using LLMs to rephrase  textual pose descriptions, which are then processed by
PoseScript \cite{posescript} to generate poses.
             Top 5, 10, 20 recall rates are reported.
 }
  \vspace{-5mm}
	\label{tab:exp_pose_generation}
\end{table}

%% file: tables/shape_HOI.tex
\begin{table}[!t]
    \centering    
    \begin{subtable}{.51\linewidth}
      \centering
      \setlength{\tabcolsep}{.6pt}
	\renewcommand{\arraystretch}{1.3}
	\renewcommand{\captionlabelfont}{\footnotesize}
        \resizebox{\linewidth}{!}{
            \makeatletter\def\@captype{table}\makeatother
            \begin{tabular}{lcccc}
                    \text{Method} & Height $\downarrow$ & Chest $\downarrow$ & Waist $\downarrow$ & Hip $\downarrow$  \\ \shline
                    LLaVA \cite{llava} & \textbf{6.7} & 16.5 & 22.9 & 17.6  \\ 
                    CLIFF-BEDLAM \cite{bedlam} & 7.8 & 8.6 & 13.5 & 7.0  \\ \rowcolor{Gray}
                    \methodname & \textbf{6.7} & \textbf{6.1} & \textbf{13.0} & \textbf{6.4}  \\ 
                \end{tabular}
            }
        \small \caption{ Comparison of body shape measurement.
                Measurement errors (in cm) on HBW \cite{shapy} are reported.} 

    \end{subtable}
    \hfill
    \begin{subtable}{.45\linewidth}
      \centering
      \setlength{\tabcolsep}{.6pt}
	\renewcommand{\arraystretch}{1.25}
	\renewcommand{\captionlabelfont}{\footnotesize}
        \resizebox{\linewidth}{!}{
        \makeatletter\def\@captype{table}\makeatother
        \begin{tabular}{lccc}
        \specialrule{0em}{-12pt}{0pt}
                \text{Method} & Precision $\uparrow$ & Recall $\uparrow$ & F1 Score $\uparrow$ \\ \shline
                LLaVA \cite{llava} & 0.26 & \textbf{0.81} & 0.39  \\ 
                GPT-4 \cite{gpt4} & 0.61 & 0.48 & 0.49  \\\rowcolor{Gray} 
                \methodname & \textbf{0.67} & 0.67 & \textbf{0.63}  \\ 
            \end{tabular}
        }
        \caption{ Comparison of HOI estimation. Precision, Recall, and F1 score on DECO \cite{deco} are reported.} 
    \end{subtable}
    \vspace{-0.2cm}
    \caption{Comparison of body shape and HOI estimation.}\label{table:shape_estimation_hoi}
    \vspace{-5mm}
\end{table}

%% file: tables/tool_ablation.tex
\begin{table*}[!t]
    \centering    
    % First Subtable
    \begin{subtable}{.50\textwidth}
        \centering
        \setlength{\tabcolsep}{3pt}
        \renewcommand{\arraystretch}{1.3}
        \resizebox{\linewidth}{!}{
            \begin{tabular}{cc|ccccc|ccccc}
                & & \multicolumn{5}{c|}{Seen Tools} & \multicolumn{5}{c}{Unseen Tools} \\
                {Paper} & {RAG} & $\text{SR}_\text{t}$ & $\text{SR}_\text{act}$ & $\text{SR}_\text{args}$ & $\text{SR}$ & IoU & $\text{SR}_\text{t}$ & $\text{SR}_\text{act}$ & $\text{SR}_\text{args}$ & $\text{SR}$ & IoU \\
                \hline
                $\times$ & $\times$ & 0.998 & 0.967 & 0.928 & 0.960 & 0.964 & 0.946 & 0.894 & 0.775 & 0.822 & 0.872 \\
                $\times$ & $\checkmark$ & \textbf{1.000} & 0.967 & 0.928 & 0.961 & 0.965 & 0.996 & 0.945 & 0.842 & 0.891 & 0.927 \\
                \checkmark & \checkmark & \textbf{1.000} & \textbf{0.974} & \textbf{0.950} & \textbf{0.970} & \textbf{0.975} & \textbf{0.999} & \textbf{0.967} & \textbf{0.893} & \textbf{0.954} & \textbf{0.953} \\
            \end{tabular}
        }
        \caption{ Ablation study of paper-based RAG mechanism.}
    \end{subtable}%
    % \hfill
    \hspace{30pt}
    % Second Subtable
    \begin{subtable}{.33\textwidth}
        \centering
        \setlength{\tabcolsep}{3pt}
        \renewcommand{\arraystretch}{1.25}
        \resizebox{\linewidth}{!}{
            \begin{tabular}{c|ccccc}
                ~~{Tool Numer}~~ & $~~\text{SR}_\text{t}~~$ & $~~\text{SR}_\text{act}~~$ & $~~\text{SR}_\text{args}~~$ & $~~\text{SR}~~$ & IoU \\
                \hline
                2 & 0.997 & 0.960 & 0.943 & 0.928 & 0.973 \\
                3 & 0.998 & 0.959 & 0.931 & 0.932 & 0.974 \\
                4 & 0.998 & 0.943 & 0.928 & 0.875 & 0.968 \\
                5 & 0.996 & 0.929 & 0.899 & 0.847 & 0.950 \\
            \end{tabular}
        }
        \caption{ Ablation study of tool number.}
        % number of tools
    \end{subtable}
    \vspace{-0.3cm}
    \caption{Ablations related to tool usage. Successful rate of thought, action, arguments, execution, and IoU are reported.}
    \label{table:tool_ablation}
    \vspace{-3mm}
\end{table*}

%% file: tables/07_feedback_ablation.tex
\begin{table}[t]
    \centering    
    \begin{subtable}{.230\textwidth}
      \centering
      \setlength{\tabcolsep}{1.25pt}
	\renewcommand{\arraystretch}{1.25}
	\renewcommand{\captionlabelfont}{\footnotesize}
        \label{subtab:hoi}
        \resizebox{\linewidth}{!}{%
        \begin{tabular}{lccc}
            Method & Precision $\uparrow$ & Recall $\uparrow$ & F1 Score $\uparrow$ \\ 
            \shline
            w/o Tool & 0.26 & \textbf{0.81} & 0.39 \\ 
            w/ Tool & \textbf{0.67} & 0.67 & \textbf{0.63}  \\
        \end{tabular}%
        }
        \caption{\footnotesize HOI Contact Detection.} 

    \end{subtable}
    \hfill
    \begin{subtable}{.235\textwidth}
      \centering
      \setlength{\tabcolsep}{1.25pt}
	\renewcommand{\arraystretch}{1.25}
	\renewcommand{\captionlabelfont}{\footnotesize}
        \label{subtab:shape_estimation_ablation}
        \resizebox{\linewidth}{!}{%
        \begin{tabular}{lcccc}
            Method & Height $\downarrow$  & Chest $\downarrow$ & Waist $\downarrow$ & Hip $\downarrow$ \\ 
            \shline
            w/o Tool & \textbf{6.7}  & 16.5 & 22.9 & 17.6  \\
            w/ Tool & \textbf{6.7}  & \textbf{6.1} & \textbf{13.0} & \textbf{6.4}  \\
        \end{tabular}%
        }
        \caption{\footnotesize Body Shape Measurement.} 
    \end{subtable}
    \vspace{-0.3cm}
    \caption{Ablation study on the impact of tool usage for human-object contact detection and body shape estimation.} 
\label{table:ablation_tool2agent}
\vspace{-4mm}\label{table:ablation_tool2agent}
\end{table}

%% file: tables/08_discrimator_ablation.tex
\begin{table}[!t]
    \centering    
    \begin{subtable}{.50\linewidth}
      \centering
      \setlength{\tabcolsep}{.75pt}
	\renewcommand{\arraystretch}{1.25}
	\renewcommand{\captionlabelfont}{\footnotesize}
        \label{subtab:mixpose}
        \resizebox{\linewidth}{!}{%
        \begin{tabular}{lccc}
            Method & MPJPE $\downarrow$ & PA-MPJPE $\downarrow$ & PA-MPVPE $\downarrow$ \\ 
            \shline
            Tool A & 126.2 & 81.4 & 101.9 \\ 
            Tool B & 124.0 & 84.6 & 104.7  \\\rowcolor{Gray}
            \methodname & \textbf{119.6} & \textbf{78.2} & \textbf{98.3}  \\
        \end{tabular}%
        }
        \small \caption{ Mesh Error (in mm) on MixPose.} 

    \end{subtable}
    \hfill
    \begin{subtable}{.47\linewidth}
      \centering
      \setlength{\tabcolsep}{.75pt}
	\renewcommand{\arraystretch}{1.25}
	\renewcommand{\captionlabelfont}{\footnotesize}
        \label{subtab:shape_estimation}
        \resizebox{\linewidth}{!}{%
        \begin{tabular}{lcccc}
            Method & Height $\downarrow$  & Chest $\downarrow$ & Waist $\downarrow$ & Hip $\downarrow$ \\ \shline
            Tool \cite{bedlam} & 7.8 & 8.6 & 13.5 & 7.0  \\ \rowcolor{Gray}
            \methodname & \textbf{6.7} & \textbf{6.1} & \textbf{13.0} & \textbf{6.4}  \\ 
        \end{tabular}%
        }
        \caption{Body Shape Measurement Error (in cm) on HBW \cite{shapy}.} 
    \end{subtable}
    \vspace{-0.3cm}
    \caption{ Study revealing how tool use improves human understanding on pose estimation and body shape measurement tasks.} 
    \vspace{-3mm}
\label{table:ablation_agent2tool}
\end{table}

%% file: sections/5_conclusion.tex
% \begin{figure*}[!t]
%   \centering  
%   \includegraphics[width=0.98\textwidth]{figures/human_interaction.pdf}
%   \vspace{-2.2mm}
%   \caption{\footnotesize Human interaction can improve performance and tool usage accuracy.}
%   % \vspace{-4mm}
%   \label{fig:human_interaction}
% \end{figure*}

\vspace{-1mm}
% \vspace{.5mm}
\section{Discussion and Concluding Remarks} 
\vspace{-1mm}
We introduce \model, an LLM-based model designed to learn the use of tools related to 3D humans and assist users in solving tasks associated with 3D humans. \model processes requests from users, analyzes the needs, and utilizes the necessary tools. It then evaluates %and synthesizes 
the tools' outputs to respond to the user's queries.% and address the problems effectively. 
% Having computers better understand 3D humans has many potential benefits in healthcare, fitness, human-robot interaction, entertainment, etc. 
% Possible negative impacts include unwanted surveillance, body shaming, and deep fakes.
% The risk-benefit ratio should be monitored and, like other LLMs, guardrails may be needed to prevent malicious use cases.

\vspace{1.5mm}
\noindent\textbf{Limitations}.
\model may  fail in certain calling scenarios, particularly when the user request is vague, and subsequent LLM internal analysis cannot rectify an incorrect initial function call. However, further interaction with users can remedy this if they provide additional information. 
Fig.~\ref{fig:human_interaction} illustrates an instance of using body estimation and face reconstruction tools for avatar creation. 
Even with the application and analysis of the tool, outcomes like height estimation may not be entirely precise. One contributing factor is the accuracy of the training data; for instance, most height labels in datasets use the official height of models or celebrities, which may not account for variations like shoe height, such as a 7-inch heel. Incorporating more cues from users, combined with the LLM's knowledge of the world and reasoning capabilities, can enhance result accuracy, as shown in Fig.~\ref{fig:human_interaction}.  
\new{
Our system is currently limited by the academic methods used. Incorporating better academic methods will enhance model performance. Notably, adding new tools requires no additional training, allowing our method to evolve and improve as new techniques are developed.  
}
% ChatHuman mainly explores tools related to text and images, and briefly delves into human motion in \supmat. Further exploration of video and motion remains for future work. 
\begin{figure}[!t]
  \centering  
  % \vspace{-2mm}
  \includegraphics[width=0.48\textwidth]{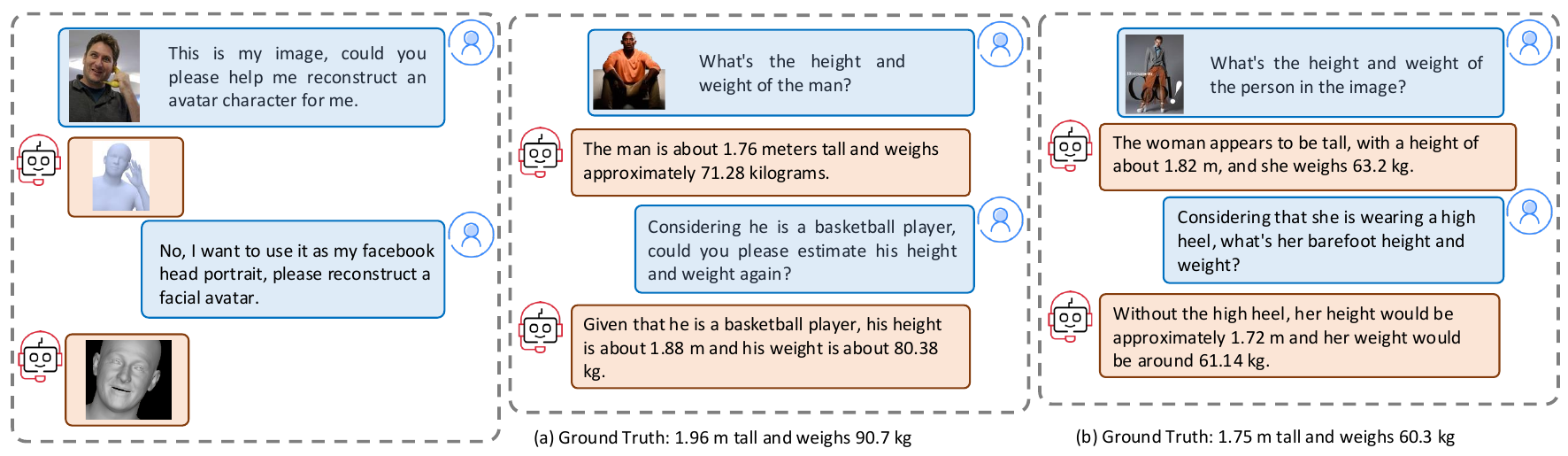}
  % \vspace{-5mm}
  \caption{Human interaction can improve the performance and tool usage accuracy of ChatHuman.}
  \vspace{-2mm}
  \label{fig:human_interaction}
\end{figure}

\vspace{1.5mm}
\noindent\textbf{Future Work}.
\model offers several avenues for future development.
In particular, user interaction/dialog offers opportunities to learn from user feedback. This could exploit reinforcement learning to refine the model's understanding.

%
% old version:
\iffalse
\model offers several avenues for future development: 
1) 
Integrated Learning and Self-Improving:
Merging tool use learning with user feedback \new{or Reinforcement Learning} to continuously refine the model's understanding and approach to 3D human tasks. 
2) User Feedback for Enhanced Training:
As shown in Fig.~\ref{fig:human_interaction}, user interaction has a tangible impact on improving the outcome. How ongoing dialogue with users might provide valuable feedback for refining the system's capabilities remains an open question. While ChatHuman focuses on 3D humans, the paradigm is general and points towards new interfaces that may open up complex vision/graphics tools to support wider applications.
\fi

%% file: sections/6_acknowledgement.tex
% \vspace{-3mm}
\section*{Acknowledgements and Disclosures}
\vspace{-1mm}
% We thank Weiyang Liu, Haiwen Feng and Longhui Yu for discussions and proofreading. 
We thank Naureen Mahmood, Nicolas Keller and 
Nicolas Heron for data collection support, and Emiliano Rallo
for video making. 
% This work was partially supported by the Max Planck ETH Center for Learning Systems.  
% CoI disclosure: \url{https://files.is.tue.mpg.de/black/CoI_CVPR_2024.txt}.
While MJB is a co-founder and Chief Scientist at Meshcapade, his research in this project was performed solely at, and funded solely by, the Max Planck Society. 

%% file: sections/X_supp.tex
\maketitlesupplementary

% \addcontentsline{toc}{section}{Supplementary Material} 
{
  \tableofcontents
}

% \newpage

As promised in the main paper, we provide additional details here on the method for integrating tool results and the model training process. This is followed by an explanation of the training data construction pipeline. We also present further details about the tool evaluation metric, prompts for body shape measurement data, and the MixPose benchmark. Furthermore, we show more ablation studies. 

\section{Additional Method Details}

\subsection{Tool Results Integration} To utilize the tool results and improve the LLMs' understanding of 3D humans, which, in turn,  enhances the LLMs' ability to apply its world knowledge to problem-solving - 
we introduce a tool-conditioned transformation  $\Psi(\cdot)$.  As shown in Figure \ref{fig:transform}, this transformation converts the varied tool outcomes $Y_m$ into textual or visual formats that the LLM can process more easily. 
For example, we transform the vertex-wise contact label predicted by DECO \cite{deco} into a body part-level description based on the vertex-to-part mapping dictionary of SMPL \cite{smpl}, and we render the mesh generated by PoseScipt \cite{posescript} into an RGB image using rendering techniques. 

\subsection{Model Training}
ChatHuman comprises a multimodal LLM $f_\phi(\cdot)$, along with a set of 3D human-related functions. During training, the tool functions are kept fixed, and only the LLM $f_\phi(\cdot)$ is finetuned using instruction-following data. Specifically, we employ LoRA~\cite{hu2021lora} with a rank of 128 and an alpha value of 256 to finetune the LLM. The trainable parameters in this setup are represented as $\phi_{lora}$. Given a user query $X_q$, the model generates a textual description of the tool invocation $Y_{tool}$ and a final textual response $Y_{t}$ after integrating the tool results. With the ground truth tool invocation labels $\hat{Y}_{tool}$ and response label $\hat{Y}_{t}$, we optimize the model using the following objective function: $\mathcal{L} = \mathbf{CE}(\hat{Y}_\text{tool}, Y_\text{tool}) + \mathbf{CE}(\hat{Y}_t, Y_t)$, where $\mathbf{CE}$ denotes the cross-entropy loss.

 % The vector similarity searching algorithm from Chroma can be found in \href{https://python.langchain.com/docs/integrations/vectorstores/chroma}{\tt LangChain/Chroma}. 
 % \TODO{more details for training}
\input{tables/A04_paper_prompt}
\subsection{Training Data Construction} 
\textbf{Tool Usage Instruction-following Data.} To teach the LLM-based agent to correctly use tools, we construct 90K instruction-response pairs about tool usage. Following GPT4Tools \cite{gpt4tools}, we provide GPT-4 \cite{gpt4} with a textual description of an image from the COCO training set \cite{coco} and a tool-related prompt containing a tool description. 
One of our key observations is that human-related tools often come with an academic paper containing rich background knowledge and varied applications, which are useful for the generation of user queries covering a wide range of application scenarios. Thus, we also incorporate the paper content into GPT-4 to generate the tool usage instruction-following data. To improve efficiency, we first prompt GPT-4 to summarize the paper content, re-articulate the tool functions and enumerate 50 potential user queries for tool activation (see main paper Fig. 6 (a)). The details of the prompt are represented in Table \ref{tab:paper_prompt}. The summarized tool description and user queries are fed to GPT-4 along with the image description to generate the instruction-following data about tool usage. Table \ref{tab:tool_instruct_generate} illustrates the prompt for the second step. 

\begin{figure}
  \centering  
  \includegraphics[width=.98\linewidth]{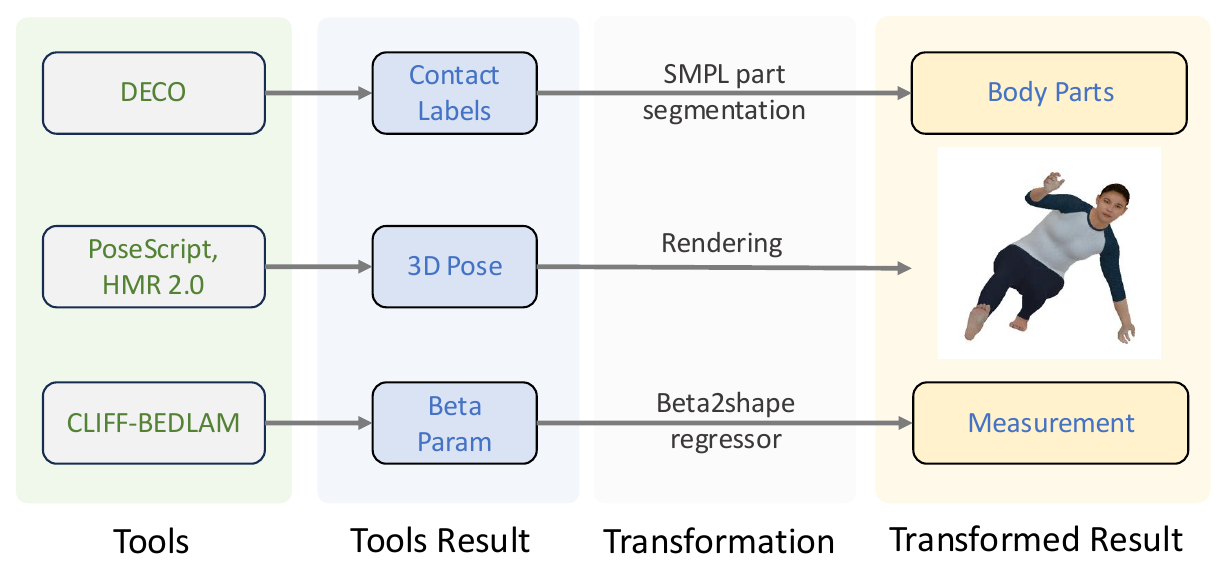}
  \vspace{-3mm}
  \caption{\footnotesize Illustration of tool-conditioned transformation process. We convert the varied tool outcomes into textual or visual formats that the LLMs can more readily process.} 
  \vspace{-3mm}
  \label{fig:transform}
\end{figure}

% \begin{figure*}[h]
%   \centering 
%   \vspace{-2mm}
%   \includegraphics[width=.99\textwidth]{figures/data_construction_v2.pdf}
%   \vspace{-3mm}
%   \caption{ Illustration of our intruction-following data construction pipeline. We construct the instruction-following data about tool usage and feedback by providing GPT-4 with various tool-related information, image content, and ground truth labels. \textcolor{gray}{Gray} text shows some example instructions.}
%   \vspace{-2mm}
%   \label{fig:data_construction}
% \end{figure*}

% \begin{figure*}[t]
%   \centering  
%   \includegraphics[width=\textwidth]{figures/pose_descrimination.pdf}
%   \vspace{-7mm}
%   \caption{\footnotesize Examples of the instruction-following data for discriminating between pose estimation results.}
%   \vspace{0mm}
%   \label{fig:pose_descrimination}
% \end{figure*}

\input{tables/R04_tool_usage_ablation}
\input{tables/A05_toolsuage_generation_prompt}
\begin{figure*}[t]
  \centering  
  \includegraphics[width=\textwidth]{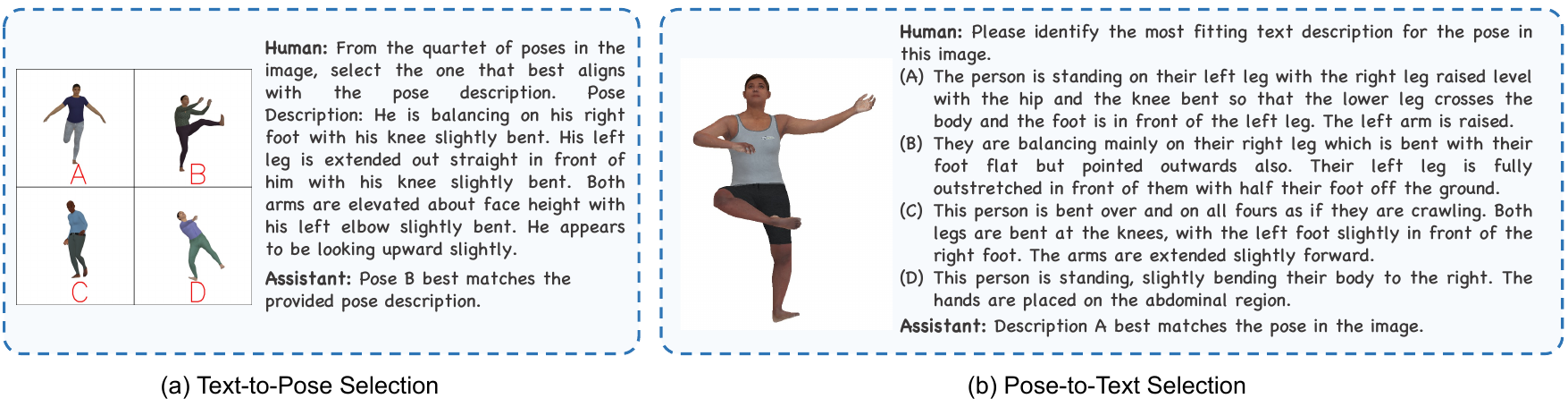}
  \vspace{-7mm}
  \caption{\footnotesize Examples of the instruction-following data for discriminating pose generation and pose description results.}
  \vspace{-2mm}
  \label{fig:pose_gen_descrimination}
\end{figure*} 

 \begin{figure*}[t]
  \centering  
  \includegraphics[width=0.75\textwidth]{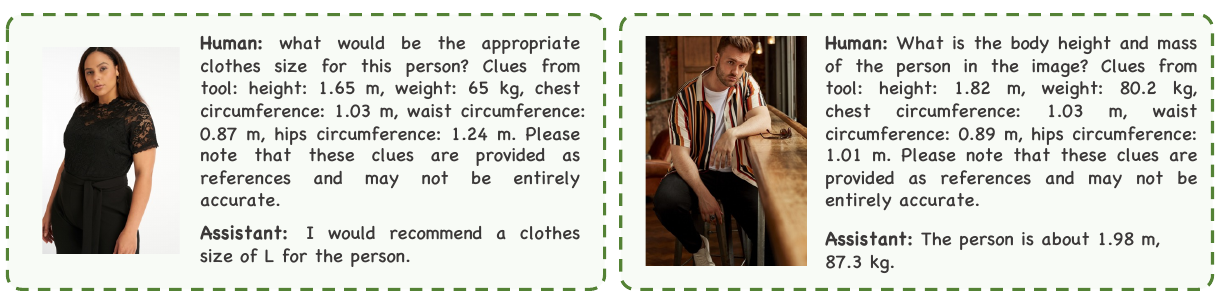}
  \vspace{-2mm}
  \caption{\footnotesize Instruction-following data for integrating results from human body estimation. Measurements of the estimated body shape from BEDLAM-CLIFF \cite{bedlam} are added to the user query as clues.}
  \label{fig:shape_integration}
\end{figure*}
\noindent\textbf{Tool Feedback Instruction-following Data.} To help the multimodal LLM model discriminate and integrate the tool results, we construct 88K pairs of instruction-following data based on existing 3D human datasets. 
\begin{itemize}
    \item \textbf{Pose Estimation Results Discrimination.} To teach the LLM-based model to discriminate the pose estimation results from different tools, we build 17K pairs of instruction-following data based on the 3DPW \cite{threedpw} and MOYO \cite{moyo} training sets. Specifically, we use HMR2.0 \cite{hmr2} and CLIFF-SMPLify \cite{cliff,smplify} to predict the human mesh and calculate the reconstruction error between the predicted mesh and ground truth mesh. Based on MPVPE, we determine which tool is better for each image and construct instruction-following data as shown in main paper Figure 6(b). Pose visualization results are rendered with Pyrender \cite{pyrender}.
    \item \textbf{Pose Generation Results Discrimination.} The human pose generation tool, PoseScript \cite{posescript}, has multiple outcomes for each text input. Here we construct 44K pairs of instruction-following data to teach the multimodal LLM-based model to discriminate the multiple pose generation results. 
    Specifically, we use PoseScript training data as the source and construct the data in two formats. The first one is about text-to-pose selection, as shown in Figure \ref{fig:pose_gen_descrimination}(a). Given a textual description, we visualize the corresponding pose and three other different poses from the training data and ask the agent to discriminate and choose the one that best aligns with the textual description. The second one is about pose-to-text matching, as shown in Figure \ref{fig:pose_gen_descrimination}(b). Given a 3D pose, we visualize it as an image by rendering the 3D body mesh in that pose.  Then, we combine it with the corresponding text description and three other pose descriptions in the format of a multiple choice question. Finally, we ask the agent to choose the one that best describes the pose shown in the image.  
    \item \textbf{Human Contact Detection Results Integration.} The outcome of the human contact prediction tool, DECO~\cite{deco}, is a vertex-wise contact prediction in a vector representation $y_c\in\mathbb{R}^{6890\times1}$, which can not be directly used as input for our multimodal LLM baseline, LLaVA. To solve this problem, we transform the vertex-wise contact label of ground-truth and DECO's result into a textual description based on the vertex-to-part mapping dictionary of the SMPL model \cite{smpl}. Subsequently, we feed the textual descriptions along with the RGB image from the DECO training set \cite{deco} into GPT-4V and prompt GPT4~\cite{gpt4} to generate instruction-following data about human-object interaction as shown in Figure \ref{fig:hoi_integration}. Notably, the transformed tool result is merged with the user query as a clue. The details of the prompt are shown in Table \ref{tab:hoi_prompt}. 
    \item \textbf{Body Shape Measurement Integration.} Similar to human contact prediction, the outcome of the body shape measurement tool is the SMPL body shape parameter $\beta\in\mathbb{R}^{10}$, which is also in a vector representation and can not be used by the LLM directly. Thus, we first convert the shape parameter into measurements based on the shape-to-measurement module from SHAPY \cite{shapy} and represent it in a textual format. Subsequently, we feed the body measurement description along with attribute labels from the SHAPY training set into GPT-4 and prompt it to generate instruction-following data about human body shape as shown in Figure \ref{fig:shape_integration}. Similarly, we merge the body measurement predicted by the tool with the user query as a clue. The prompt for GPT-4 is detailed in Table \ref{tab:shape_prompt}.

\end{itemize}
\input{tables/A07_HOI_prompt}

\section{Evaluation Metric and Benchmark Details}
\subsection{Evaluation Metric}
\textbf{Tool Usage.} We use the metrics proposed in GPT4Tools \cite{gpt4tools} to measure the tool usage accuracy, including:

\begin{itemize}
    \item \textbf{Successful Rate of Thought} ($\text{SR}_\text{t}$), which measures the decision accuracy, calculated as $\text{SR}_\text{t}=\frac{1}{N}\sum_{i=1}^{N}\mathbb{I}(\tau_i)$, where $N$ is the number of instructions and $\tau_i$ is a singular process. When the predicted thought is the same as the ground-truth thought, $\mathbb{I}(\tau_i)$ is equal to 1, and 0 otherwise.
    \item \textbf{Successful Rate of Action} ($\text{SR}_\text{act}$), which  measures the tool name prediction accuracy, calculated as $\text{SR}_\text{act}=\frac{1}{N}\sum_{i=1}^{N}\mathbb{I}(\alpha_i)$, where $\alpha_i$ is the matching process of the tool name. If the predicted tool name is correct, $\mathbb{I}(\alpha_i)$ is equal to 1, and 0 otherwise.
    \item \textbf{Successful Rate of Arguments} ($\text{SR}_\text{args}$), which measures the tool arguments prediction accuracy, calculated as: 
    \begin{equation}
        \text{SR}_\text{args} = \frac{1}{N}\sum_{i=1}^{N}\eta_{i}, ~~\eta_{i} = \frac{1}{K}\sum_{j=1}^{K}\eta_{i,j},
    \end{equation}
    where $K$ is the number of tool arguments. When the argument is a file name, $\eta_{i,j}$ equals  1 if the predicted file name is the same as the ground-truth file name, and 0 otherwise. When the argument is text, $\eta_{i,j}$ equals the BLEU score between the predicted and ground-truth text.
    \item \textbf{Intersection over Union} ($\text{IoU}$), which quantifies the percent overlap between the predicted text and ground-truth text.
 \end{itemize}

\noindent\textbf{Human Understanding.} We use the following evaluation metrics to measure the performance of \model in human-related tasks:
\begin{itemize}
    \item \textbf{Pose Estimation.} We adopt the same evaluation metrics as ChatPose \cite{feng2024chatpose} to evaluate the 3D pose estimation accuracy, including Mean Per-Joint Position Error (MPJPE), Mean Per-Joint Position Error after Procrustes alignment (PA-MPJPE), Mean Per-Joint Rotation Error (MPJRE), and Mean Per-Vertex Position Error (MPVPE).
    \item \textbf{Pose Generation.} We use the evaluation metrics established in PoseScript \cite{posescript}, including the text-to-pose recall rate $R^{P2T}$ and pose-to-text recall rate $R^{T2P}$ of the retrieval models trained on real poses and evaluated on generated poses. We use the retrieval model from the jounal-version of PoseScript \cite{posescript} and ChatPose \cite{feng2024chatpose} for the classical pose generation and speculative pose generation tasks, respectively.
\end{itemize}

\begin{figure*}[t]
  \centering  
  \includegraphics[width=0.85\textwidth]{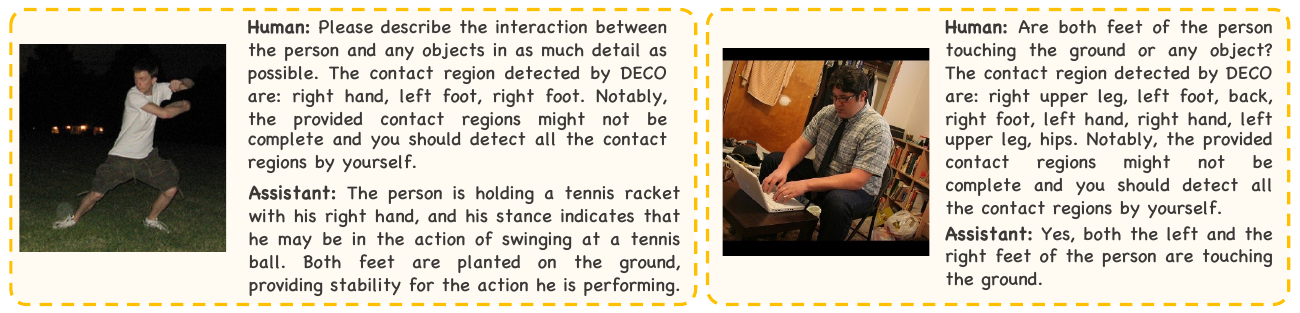}
  \vspace{-2mm}
  \caption{\footnotesize Instruction-following data about integrating results from human contact detection. The contact labels detected by DECO \cite{deco} are combined with the user query.}  
  \vspace{-2mm}
  \label{fig:hoi_integration}
\end{figure*}

\subsection{Benchmarks}
% \TODO{Add prompt for benchmarks}
\noindent\textbf{Tool Usage Benchmark.} To evaluate the tool usage accuracy of our method, we construct a validation and test set. The validation set has 1000 samples with the same tools as the training set, while the test set includes 689 samples related to 3 tools unseen during training. Similar to the training data construction, we feed a textual description of an image from the COCO validation set, a tool description, and some examples summarized from the tool paper into GPT-4 and prompt it to generate instruction-following data about tool usage. We use the image description captioned by LLaVA \cite{llava} instead of the original image captions to ensure a difference between training and test sets. Finally, we manually check the question-answering pairs to ensure the accuracy of the benchmark. 

\noindent\textbf{MixPose Benchmark.} To validate whether the multimodal LLM-based agent can discriminate the pose estimation results from different tools, we build a new benchmark, MixPose. Considering that different tools excel in different scenarios, a benchmark covering diverse scenarios and corner cases is needed. To construct this benchmark, we selected 100 images featuring extreme camera angles from  MoYo \cite{moyo} test set, 100 full-body images from the 3DPW test set, and 100 images with significant truncation also from the 3DPW test set.  This approach ensures our benchmark includes typical in-the-wild images, shots taken from extreme views, and images with heavy truncation. 
This diverse distribution allows us to test whether the agent can accurately choose the right tool based on sceneries of the image and the tool performance.  
To get the truncated images from 3DPW, we resize the human bounding box by 2/3 and crop images based on the rescaled human bounding box. 

% \noindent\textbf{Body Shape Measurement}

\new{
\section{Ethnical Statement}
\vspace{-1mm}
Our work explores utilizing tools for human-centric tasks. All 3D human data and tools used is publicly available intended for research. No identifiable or biometric data is used. 
We acknowledge the potential risks of deploying such systems in real-world applications, including surveillance, biased inferences, or non-consensual modeling. We strongly discourage such uses. We encourage future users of our system and dataset to consider fairness, inclusivity, and transparency in their work. 
Our code and data are released under licenses that restrict commercial use and require proper attribution, to encourage responsible and ethical research.
}

\section{Additional Results}

\new{\subsection{RAG Qualitative Results.} 
As mentioned in main paper Sec.~3.2, many tools require background knowledge and have various application scenarios, which can be derived from the scientific paper. Fig.~\ref{fig:rag_examples} shows some retrieved examples for the ``Body Pose Estimation" tool from our RAG Mechanism. }
\begin{figure}[h]
  \centering  
    % \vspace{-0.35cm}
  % \includegraphics[width=\textwidth]{figures/RAG_examples.pdf}
  \includegraphics[width=\linewidth]{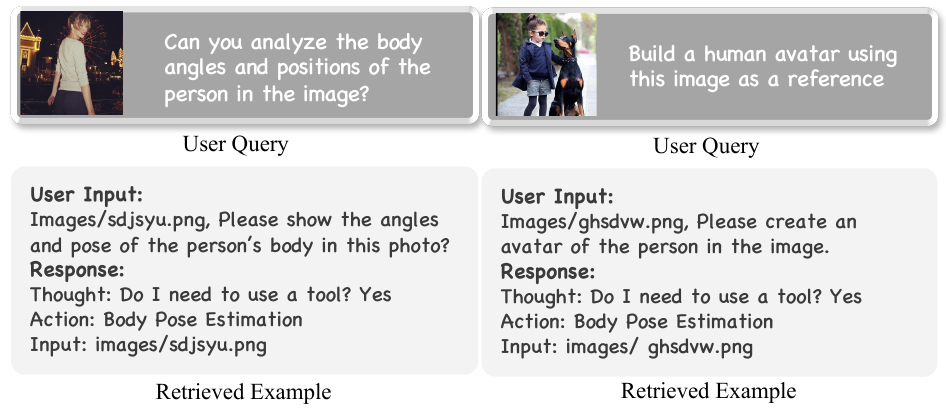}
    \vspace{-0.35cm}
  \caption{\footnotesize Qualitative examples of the scientific-paper-based RAG mechanism. A single tool can have multiple application scenerios.}
  \label{fig:rag_examples}
    \vspace{-0.36cm}
\end{figure} 

\input{tables/R01_hmr_sota}
\new{\subsection{GPT4Tool and Visual ChatGPT-4}
Tab.~\ref{tab:r_exp_hmr} shows a comparison of ChatHuman, GPT4Tools \cite{gpt4tools}, and Visual ChatGPT-4 \cite{wu2023visual-chatGPT} on the task of pose estimation, following the metrics in main paper Tab.~3.  We change the LLM agents while keeping the other setting unchanged. For GPT4Tools, we finetune it with our training data about tool usage for a fair comparison. When the method selects a wrong tool and fails to get a pose output, we calculate the error using a T-pose as the result. ChatHuman outperforms these baselines in the standard pose estimation task and the more complex reasoning-based pose estimation task. 
}

\input{tables/R02_tool_resample}
\new{
\subsection{K-fold Validation of Tool Usage}
To further verify the robustness of tool utlizaiton, we conducted K-fold validation by splitting 26 tools into 9 folds, each containing 2-3 tools. The experiment was repeated three times, each time using a random fold as the unseen tools for evaluation and the rest for training. As shown in Tab.~\ref{table:tool_kfold}, consistent performance across trials shows strong generalization to unseen tasks. Tab.~\ref{table:unseen_tool_num} shows that our method maintains robust performance as the number of unseen tools grows. 
}

\input{tables/R03_unseen_tool_num}
\new{
\subsection{GPT4 replaces LLaVA for comparison}{
We replace LLaVA  with GPT4 and evaluate performance on the tool-use benchmark. As in Tab.~\ref{table:tool_kfold}, we use K-fold validation with 3 repeats, reporting average accuracy in Tab.~\ref{tab:gpt4_rag}. We will include it in Tab.~2 of the main paper. 
}
}

\subsection{Paper Components.} To improve the tool usage accuracy, we propose a paper-based RAG mechanism. 
We conduct an ablation study to analyze the effects of each component of the paper. The baseline model is derived by removing the RAG operation and is trained with the instruction-following data constructed based on the manually defined tool descriptions and examples (T), without referring to paper content. We then add the tool descriptions and examples summarized from different paper components into the prompt and request GPT-4 to generate new instruction-following data. The paper components include the abstract (A), introduction (I), related work (R), method (M), and experiment section (E). The results are presented in Table \ref{tab:paper_ablation}. As shown, incorporating the paper content  consistently improves the accuracy of tool usage. Interestingly, feeding the abstract and introduction of the paper achieves the best performance, and adding additional paper components like the experiments does not result in further improvement. This is expected since the abstract and introduction have covered the tool function and potential applications in most cases. We also find that examples generated based on paper content can be noisy in some cases, and thus we manually check the examples and remove the incorrect samples before feeding them to GPT-4. This data cleaning operation improves the accuracy.

\input{tables/A02_paper_components}

\input{tables/A03_model_size}

\subsection{Base Model for Value Extraction.} During our body shape measurement and human contact detection experiments, the answer of \model is a sentence and thus could not be directly used to compute the evaluation metrics. Thus, we need to prompt an LLM to extract the value from the sentence and compare the extracted value with the ground truth label to calculate the metric. The prompts for body shape measurement and contact estimation are shown in Table \ref{tab:evaluation_prompt}. Here we use two different LLM models for value extraction and report the results in Table \ref{table:ablation_base_model}. As shown, the choice of the LLM model for value extraction does not introduce a significant difference.

\input{tables/A06_text_embedding}

\subsection{Text Embedding Model in Retrieval.} During the paper-based retrieval mechanism, we retrieve a relevant example by matching the text embedding of the query and those within the tool documents using a pretrained text embedding model \cite{su2022one}. Here we conduct an ablation study of the text embedding model to analyze the robustness of our RAG mechanism. We experiment with two models, i.e., instructor-xl and instructor-base. As shown in Table \ref{tab:text_embed_ablation}, the performance of our method is not greatly affected by the text embedding model, which demonstrates the robustness of \model.

\subsection{Tool Use in Multi-turn Dialogue.} In many real-world scenarios, the models need to correctly use the tool based on the context within a multi-turn dialogue. Here, we build a benchmark to evaluate the tool use accuracy within multi-turn conversations. We randomly select several single-turn question-answer pairs from the original multi-turn tool use benchmark and ask GPT-4 to merge them into a multi-turn dialogue. We then evaluate the performance of GPT4Tools~\cite{gpt4tools}, Visual ChatGPT~\cite{wu2023visual-chatGPT}, and ChatHuman on the built benchmark, which contains 1000 instruction-response pairs. As shown in Table \ref{tab:tool_usage_multiturn}, ChatHuman achieves a superior performance in the multi-turn setting, demonstrating its capacity to understand the comprehensive context information within the multi-turn dialogue and correctly use the tool to solve the problems.

% \begin{figure}[t]
%   \centering  
%   \includegraphics[width=0.48\textwidth]{figures/failure_cases.pdf}
%   % \vspace{-2mm}
%   \caption{\footnotesize Failure cases of ChatHuman. When the user inputs a vague query or requests a task that can not be finished by current tools, ChatHuman might fail.}
%   \label{fig:failure_cases}
%   \vspace{-2mm}
% \end{figure}
\input{tables/A09_evaluation_prompt}
\input{tables/A10_multiturn_tool_use}
\input{tables/A08_shape_prompt}

\subsection{Tool Graph Construction.}
\new{We prompt GPT-4 to construct a tool graph with three structure types: nodes (single tool calls for simple tasks), chains (tool sequences for dependent tasks), and directed acyclic graphs (DAGs) \cite{shen2023taskbench} for complex multi-branch operations. Table \ref{tab:tool_graph_construction} shows the detail of the prompt.}

\input{tables/R05_tool_graph_construction}

% \input{sections/appendixsec/6_results}

% \subsection{Failure Cases and Limitations.} There are mainly two kinds of failure cases of \model. Firstly, ChatHuman may fail in certain calling scenarios, particularly when the user request is vague, as also shown in Figure 7 in the main paper. Secondly, when the user query requires a task (e.g., general object video generation) that is not supported by the tools in the tool libraries, \model might incorrectly choose a tool to generate a response, instead of informing the user that the task cannot be completed and recommending the creation or introduction of a new tool. As shown in Figure~\ref{fig:failure_cases}(b), there is a need for tools for video generation. Future developments may include expanding to tools related to these scenarios. 

% \input{sections/appendixsec/3_training_data}
% \input{sections/appendixsec/1_tool}
% \input{sections/appendixsec/2_evaluation}
% \input{sections/appendixsec/4_benchmark}
% \input{sections/appendixsec/5_ablation}

%% file: tables/A04_paper_prompt.tex
% \begin{table}[t]
%     \centering
%     \footnotesize
%     \begin{tcolorbox} 
%     \renewcommand\arraystretch{1.2} %
%     \begin{tabular}{p{\linewidth}}
\begin{table}[t]
    \centering
    \footnotesize
    \begin{tcolorbox} 
    \renewcommand\arraystretch{1.2} %
    \begin{tabular}{p{\textwidth}}
        You are an AI visual assistant tasked with analyzing a paper on a method in the field of 3D human modeling. Your goal is to extract key information about the method—its name, purpose, uses, and potential application scenarios. Based on this, you need to succinctly define the method in the following formats ``Method name is a tool to do something. Useful when you want to do something. Like: user query.'' \\  \\
        Subsequently, craft 50 diverse, realistic user prompts that indirectly pertain to using this method. These queries should be framed as questions, demands, or scenarios from consumers who are unaware of the method's name but whose needs align with its capabilities. Assume that consumers have an image and seek assistance in achieving a task related to the image using this method. Each prompt should introduce the task of the user in an imperative tone. The prompt should specify and refer to the image.  \\ \\
        
        Here is one example: \\   \\
         
        Method definition: \\
        name=``HMR2.'', \\
        description=``HMR2.0 is a tool to estimate the 3D pose and shape of the person in the image. Useful when you want to detect poses of the humans in the image. Like: estimate the human poses in the image.''  \\ \\
        
        Possible queries: \\
        1. Can you help me estimate the pose of the person in the photo? \\
        2. Please reconstruct a 3D avatar for the person in the image. \\
        3. Could you please estimate the SMPL parameter of the man in the picture? \\
        4. Someone in the image is riding a bicycle, could you please help me estimate her pose?
    \end{tabular}
    \end{tcolorbox} 
    \vspace{-2mm}
    \caption{\footnotesize
        Prompt to request GPT-4 to summarize paper content, rearticulate tool functions, and enumerate possible user queries for tool activation.
    }
    \label{tab:paper_prompt}
    \end{table}

%% file: tables/R04_tool_usage_ablation.tex
\begin{table}[h]
	\centering
	\tiny
 \vspace{-3.5mm}
	\setlength{\tabcolsep}{1.2pt}
	\renewcommand{\arraystretch}{1.45}
	\renewcommand{\captionlabelfont}{\footnotesize}
 %\vspace{-.25mm}
	\resizebox{\columnwidth}{!}{
		\begin{tabular}{l|ccccc|ccccc}
			&  \multicolumn{5}{c|}{ Seen Tools} & \multicolumn{5}{c}{Unseen Tools}\\
            % \hline
			{Method} & $~\text{SR}_\text{t}~$ & $~\text{SR}_\text{act}~$ & $~\text{SR}_\text{args}~$ & $~\text{SR}~$ &~IoU~& $~\text{SR}_\text{t}~$ & $~\text{SR}_\text{act}~$ & $~\text{SR}_\text{args}~$ & $~\text{SR}~$ & ~IoU~ \\
           % Visual ChatGPT-4 \cite{wu2023visual-chatGPT}  &  0.892 &     0.802     &     0.715   &  0.753 &    0.797 & 0.998 &     0.913   &  0.801 &    0.872 & 0.907    \\
           \hline
           Ours w/ GPT-4  &  0.953 &     0.920    &     0.732   &  0.751 &    0.875 & 0.969 &     0.924   &  0.734 &    0.746 & 0.876    \\
			ChatHuman  &  \bf 1.000 &    \bf  0.973     &    \bf  0.951   & \bf  0.966 &   \bf  0.974 & \bf 1.000 &    \bf  0.968  & \bf  0.893 &   \bf  0.953 &\bf  0.953    \\
		\end{tabular} 
	}
 \vspace{-0.4cm}
    \caption{\footnotesize{Comparison of tool use accuracy.}
	}
 % \vspace{-3.5mm}
	\label{tab:gpt4_rag}
\end{table}

%% file: tables/A05_toolsuage_generation_prompt.tex
\begin{table}[t]
    \centering
    \footnotesize
    \begin{tcolorbox} 
    \renewcommand\arraystretch{1.2} %
    \begin{tabular}{p{\textwidth}}
    
    Given an image whose image path is ``example.jpg''. Image caption: ``\{caption\}''. 
The image caption includes detail image description and each object paired with the bounding box [x1, y1, x2, y2]. For the bounding box, (x1, y1) refers to the top left, and (x2, y2) refers to the bottom right. x1 is less than x2, and y1 is less than y2.  \\ \\

Below are 26 visual tools. Each tool is defined as ``tool name: usage scenario, and tool arguments''. \\ \\

Please generate 3 visual instructions for each tool, so you need to generate 66 visual instructions in total. \\ 
The generated instructions should follow the format of ``instruction content, [tool name, tool arguments]''.  Each instruction must relate to the caption and can be solved by the tool.  \\
You can not revise the ``tool name'', or add any other fake tools that are not defined. You must keep the correct ``tool arguments''. \\ \\

Tools:
\{tool description\} \\ \\

Note that your generated visual instructions should be highly related to the image caption. Directly reply to me with the list, here are some examples:
\{examples\} \\ \\

Diversify the instructions to cover a wide range of possible user queries.  Feel free to adapt and rephrase the examples provided to generate diverse, complex, and deceptive instructions as much as possible.  \\
For example, you can also change the subject position or the person and pose description positions. Don't use too much imperative sentence, you should also use interrogative sentences.

    \end{tabular}
    \end{tcolorbox} 
    \caption{\footnotesize
        Prompt to request GPT-4 to generate instruction-following data about tool usage based on the image description, tool description, and tool usage examples summarized from tool paper.
    }
    \label{tab:tool_instruct_generate}
    \end{table}

%% file: tables/A07_HOI_prompt.tex
\begin{table*}[t]
    \centering
    \footnotesize
    \begin{tcolorbox} 
    \renewcommand\arraystretch{1.2} %
    \begin{tabular}{p{\textwidth}}
        You are an AI visual assistant, and you are seeing a single image and a sentence about the human-object contact regions of the person in the image.
The sentence include the human-object contact body parts of the person. Notably, the provided contact regions might not be complete and you should detect all the contact regions by yourself.   \\ \\ 

Design a conversation between you and a person asking about the human-object contact information of the person. The answers should be in a tone that a visual AI assistant is seeing the image and answering the question.
Ask diverse questions and give corresponding answers.  \\ \\

Include questions asking about the person's human-object contact information, etc. Only include questions that have definite answers:
(1) one can see the content in the image that the question asks about and can answer confidently;
(2) one can determine confidently from the image that it is not in the image.
Do not ask any questions that cannot be answered confidently. \\ \\

Provide detailed answers when answering complex questions. In your answer, you should imitate as if you see the image and the contact regions are estimated by you.
You should only ask questions about the human-object interaction. The answer should be as detailed as possible. Don't mention any other irrelevant information!
Directly reply to me with a list, here are some examples: \\
1. Please help me detect the contact regions of the person in the image. [The person's contact region includes his feet, hands, and back. His feet touch the ground and his hands are holding a mobile phone.] \\
2. Does the person's hand contact any objects? [Yes, her right hand holds an umbrella.] \\
3. Describe the human-object interaction information of the person, as detailed as possible. [The person holds a phone with his left hand and stands on a skateboard with both their left and right feet.] \\ \\

Notably, at least one question is to ask all the contact regions of the person.
In your answer, you should distinguish and specify the left and right body parts. Notably, you should distinguish based on the body pose and orientation. If the person is facing the camera, the hand, foot, and ear on the left side of the image is the person's right hand, right foot, and right ear, and the one on the right side of the image is the person's left hand, left foot, and left ear. If the person has their back to the camera, the one on the left side of the image is the person's left body part, and the one on the right side of the image is the person's right body part.

    \end{tabular}
    \end{tcolorbox} 
    \caption{\footnotesize
        Prompt to request GPT-4V to generate instruction-following data about human-object interaction based on the  contact description and RGB image.
    }
    \label{tab:hoi_prompt}
    \end{table*}

%% file: tables/R01_hmr_sota.tex
\begin{table}[h]
	\centering
	\scriptsize
 \vspace{-0.3cm}
        \setlength{\tabcolsep}{1.2pt}
	\renewcommand{\arraystretch}{1.25}
	\renewcommand{\captionlabelfont}{\footnotesize}
	\resizebox{\columnwidth}{!}{
		\begin{tabular}{l|ccc|ccc}
			&  \multicolumn{3}{c|}{ 3DPW} & \multicolumn{3}{c}{RPE Benchmark}\\
			{Method} & {MPJPE $\downarrow$} & {PA-MPJPE $\downarrow$} & {MPJRE $\downarrow$} & {MPJPE $\downarrow$} & {PA-MPJPE $\downarrow$} & {MPJRE $\downarrow$} \\
                \hline

                GPT4Tools-FT      &   114.1      &    71.0     &  10.2 &   190.5     &  100.6     &  11.6 \\
                Visual ChatGPT-4      &    103.8     &    63.1      &   10.1  &   168.4  & 82.3 & 10.7  \\
                ChatHuman     &    \textbf{91.3}       &     \textbf{58.7}    &   \textbf{9.2}    &      \textbf{147.2}     & \textbf{79.1} & \textbf{10.3}    \\
		\end{tabular} 
	}
 %\vspace{0.75mm}
 \vspace{-0.35cm}
    \caption{
	           \footnotesize{Comparison of vanilla human pose estimation and reasoning-based pose estimation on 3DPW and RPE.}
	}
	\label{tab:r_exp_hmr}
 \vspace{-3.6mm}
\end{table}

%% file: tables/R02_tool_resample.tex
\begin{table}[h]
    % \vspace{-0.3cm}
        \centering
        \tiny 
        \setlength{\tabcolsep}{3pt}
        \renewcommand{\arraystretch}{1.25}
        \resizebox{0.75\linewidth}{!}{
            \begin{tabular}{c|ccccc}
                ~~~~ & $~~\text{SR}_\text{t}~~$ & $~~\text{SR}_\text{act}~~$ & $~~\text{SR}_\text{args}~~$ & $~~\text{SR}~~$ & IoU \\
                \hline
                Trial 1 & 0.999 & 0.967 & 0.893 & 0.954 & 0.953 \\
                Trial 2 & 1.000 & 0.971 & 0.895 & 0.955 & 0.954 \\
                Trial 3 & 1.000 & 0.965 & 0.890 & 0.951 & 0.951 \\
                \hline
                Average & 1.000 & 0.968 & 0.893 & 0.953 & 0.953 \\
            \end{tabular}
        }
    \vspace{-0.32cm}
    \caption{\footnotesize{K-fold cross-validation with 3 trials.}}
    \label{table:tool_kfold}
    \vspace{-0.3cm}
    
\end{table}

%% file: tables/R03_unseen_tool_num.tex
\begin{table}[h]
    \vspace{-0.3cm}
        \centering
        \tiny
        \setlength{\tabcolsep}{3pt}
        \renewcommand{\arraystretch}{1.25}
        \resizebox{0.78\linewidth}{!}{
            \begin{tabular}{c|ccccc}
                Unseen Tool & $~~\text{SR}_\text{t}~~$ & $~~\text{SR}_\text{act}~~$ & $~~\text{SR}_\text{args}~~$ & $~~\text{SR}~~$ & IoU \\
                \hline
                3 & 1.000 & 0.965 & 0.890 & 0.951 & 0.951 \\
                5 & 1.000 & 0.962 & 0.885 & 0.939 & 0.946 \\
                7 & 0.999 & 0.948 & 0.881 & 0.929 & 0.944 \\
            \end{tabular}
        }
    \vspace{-0.3cm}
    \caption{\footnotesize{Accuracy on more unseen tools.}}
    % Ablation study of the number of unseen tools. Accuracies on unseen tools are reported.}}
    \label{table:unseen_tool_num}
    \vspace{-0.3cm}
    
\end{table}

%% file: tables/A02_paper_components.tex
\begin{table}[t]
	\centering
	\scriptsize
        \setlength{\tabcolsep}{1.75pt}
	\renewcommand{\arraystretch}{1.25}
	\renewcommand{\captionlabelfont}{\footnotesize}
	\resizebox{\columnwidth}{!}{
		\begin{tabular}{c|ccccc|c|ccccc|ccccc}
			& \multicolumn{5}{c|}{Paper} & & \multicolumn{5}{c|}{ Seen Tools} & \multicolumn{5}{c}{Unseen Tools}\\
			T & A & I & R & M & E & C & $~~\text{SR}_\text{t}~~$ & $~~\text{SR}_\text{act}~~$ & $~~\text{SR}_\text{args}~~$ & $~~\text{SR}~~$ &~~IoU~~& $~~\text{SR}_\text{t}~~$ & $~~\text{SR}_\text{act}~~$ & $~~\text{SR}_\text{args}~~$ & $~~\text{SR}~~$ & ~~IoU~~ \\
			\shline
    \checkmark & & & & & & \checkmark & 1.0& 0.97& 0.93 & 0.96 & 0.96 & 0.95 & 0.89 & 0.78 & 0.82 & 0.87 \\
	  \checkmark & \checkmark & & & & & & 1.0& 0.97& 0.95 & 0.97 & 0.97 & 0.99 & 0.94 & 0.85 & 0.90 & 0.93 \\
	  \checkmark & \checkmark & \checkmark & & & & &1.0 &0.97 &0.95 & 0.97 &0.97& 1.0&0.97&0.86&0.91&0.94 \\
	  \checkmark & \checkmark & \checkmark & \checkmark & & & & 1.0 & 0.98 & 0.95& 0.97& 0.97& 1.0 & 0.97 & 0.84 &0.91 & 0.93 \\
	  \checkmark & \checkmark & \checkmark & \checkmark & \checkmark & & &  1.0 & 0.98 & 0.94 & 0.97 & 0.97 & 0.99 & 0.95 & 0.82 & 0.87 &0.92 \\
	  \checkmark & \checkmark & \checkmark & \checkmark & \checkmark & \checkmark & &1.0 &0.98 & 0.95 & 0.97 &0.97 &1.0 &0.96 & 0.86 &0.91 & 0.94 \\
	  \checkmark & \checkmark & \checkmark & &  &  & \checkmark & 1.0 & 0.97& 0.95&0.97 & 0.98 &1.0 & 0.97 & 0.89 & 0.95 & 0.95 \\
		\end{tabular} 
	}
 % \vspace{2mm}
    \small\caption{\footnotesize
	            Ablation study on the impact of each paper component in the paper-based RAG mechanism. T denotes tool description, A, I, R, M, E are abstract, introduction, related work, method, experiment section from the paper, and C denotes the data after manually cleaning. Successful rate of thought ($\text{SR}_\text{t}$), action ($\text{SR}_\text{act}$), arguments ($\text{SR}_\text{args}$), execution ($\text{SR}$), and IoU are reported. 
	}
	\label{tab:paper_ablation}
    % \vspace{-2mm}
\end{table}

%% file: tables/A03_model_size.tex
\begin{table}[t]
    \centering    
    \begin{subtable}{.44\linewidth}
      \centering
      \setlength{\tabcolsep}{1.25pt}
	\renewcommand{\arraystretch}{1.25}
	\renewcommand{\captionlabelfont}{\footnotesize}
        % \label{subtab:hoi}
        \resizebox{\linewidth}{!}{%
        \begin{tabular}{lccc}
            Evaluator & ~Precision $\uparrow$~ & ~Recall $\uparrow$~ & ~F1 Score $\uparrow$~ \\ 
            \shline
            GPT-3.5~ & 0.67 & 0.67 & 0.63 \\ 
            GPT-4~ & 0.69 & 0.69 & 0.64  \\
        \end{tabular}%
        }
        \caption{\footnotesize HOI Contact Detection.} 

    \end{subtable}
    \hfill
    \begin{subtable}{.53\linewidth}
      \centering
      \setlength{\tabcolsep}{1.25pt}
	\renewcommand{\arraystretch}{1.25}
	\renewcommand{\captionlabelfont}{\footnotesize}
        % \label{subtab:shape_estimation_ablation}
        \resizebox{\linewidth}{!}{%
        \begin{tabular}{lccccc}
            Evaluator & ~Height $\downarrow$~ & ~Weight $\downarrow$~ & ~Chest $\downarrow$~ & ~Waist $\downarrow$~ & ~Hip $\downarrow$~ \\ 
            \shline
            GPT-3.5 & 6.7 & 10.4  & 6.1 & 13.0 & 6.4  \\
            GPT-4 & 6.7 & 10.4 & 6.1 & 13.0 & 6.4  \\
        \end{tabular}%
        }
        \caption{\footnotesize Body Shape Measurement.} 
    \end{subtable}
    \vspace{-0.25cm}
    \small\caption{\footnotesize Ablation study on the base model for value extraction.} 
\label{table:ablation_base_model}
% \vspace{-7mm}
\end{table}

%% file: tables/A06_text_embedding.tex
\begin{table}[t]
	\centering
	\scriptsize
        \setlength{\tabcolsep}{1.75pt}
	\renewcommand{\arraystretch}{1.25}
	\renewcommand{\captionlabelfont}{\footnotesize}
	\resizebox{\columnwidth}{!}{
		\begin{tabular}{c|ccccc|ccccc}
			& \multicolumn{5}{c|}{ Seen Tools} & \multicolumn{5}{c}{Unseen Tools}\\
			{Method} & $~~\text{SR}_\text{t}~~$ & $~~\text{SR}_\text{act}~~$ & $~~\text{SR}_\text{args}~~$ & $~~\text{SR}~~$ &~~IoU~~& $~~\text{SR}_\text{t}~~$ & $~~\text{SR}_\text{act}~~$ & $~~\text{SR}_\text{args}~~$ & $~~\text{SR}~~$ & ~~IoU~~ \\
			\shline
       instructor-base             &     1.000     &     \textbf{0.975}   &  0.947 &    \textbf{0.972} & 0.974 &     0.997   &  0.950 &    0.884 & 0.949 & 0.949   \\
	instructor-xl	  &  \textbf{1.000} &     0.974     &     \textbf{0.950}   &  0.970 &    \textbf{0.975} & \textbf{0.999} &     \textbf{0.967}   &  \textbf{0.893} &    \textbf{0.954} & \textbf{0.953}    \\
		\end{tabular} 
	}
 % \vspace{0.7mm}
    \small\caption{\footnotesize
	            Ablation study of text embedding model for RAG. Successful rate of thought ($\text{SR}_\text{t}$), action ($\text{SR}_\text{act}$), arguments ($\text{SR}_\text{args}$), execution ($\text{SR}$), and IoU are reported. 
	}
	\label{tab:text_embed_ablation}
 \vspace{-4mm}
\end{table}

%% file: tables/A09_evaluation_prompt.tex
% \begin{table}
%     \centering
%     \footnotesize
%     \begin{tcolorbox} 
%     \renewcommand\arraystretch{1.2} %
%     \begin{tabular}{p{\textwidth}}
\begin{table}[t]
    \centering
    \footnotesize
    \begin{tcolorbox} 
    \renewcommand\arraystretch{1.2} %
    \begin{tabular}{p{\textwidth}}
    (a) You are an AI assistant. Your input will be a description of body measurements, including height, weight, chest circumference, hip chest circumference, and waist circumference. Your task is to extract the value of each attribute and return a result like: \\ 
``height: 1 m, weight: 1 kg, chest circumference: 1 m, waist circumference: 1 m, hip circumference: 1 m'' \\

If there is no measurement value, return ``There is no measurement value.'' \\ \\

(b) You are an AI assistant. Your input will be a description of the human-object interaction information of a person.  Your task is to extract the body parts that contact with objects and return a list. Consider the following possible body parts: right hand, right upper leg, left arm, left leg, left foot, back, left shoulder, right shoulder, right foot, head, right arm, left hand, right leg, left forearm, right forearm, neck, left upper leg, hips.
    \end{tabular}
    \end{tcolorbox} 
    \vspace{-3mm}
    \caption{\footnotesize Prompt to extract the target values from a sentence generated by \model for metric computation. Prompt (a) is used to request GPT-3.5 for body measurement values. Prompt (b) instructs GPT-3.5 to extract the body part names. }
    \label{tab:evaluation_prompt}
    \vspace{-2mm}
    \end{table}

%% file: tables/A10_multiturn_tool_use.tex
\begin{table}[t]
	\centering
	\scriptsize
	\setlength{\tabcolsep}{5pt}
	\renewcommand{\arraystretch}{1.45}
	\renewcommand{\captionlabelfont}{\footnotesize}
 \vspace{-.25mm}
	\resizebox{0.99\columnwidth}{!}{
		\begin{tabular}{l|ccccc}
&  $~~\text{SR}_\text{args}~~$  & $~~\text{SR}~~$ &~~IoU~~& $~~\text{SR}_\text{t}~~$ & $~~\text{SR}_\text{act}~~$  \\
			\shline
            GPT4Tools \cite{yang2023gpt4tools}  &  0.582 &  0.551        &  0.553    & 0.513  &   0.612    \\
			Visual ChatGPT-3.5 \cite{wu2023visual-chatGPT} & 0.438  &   0.203      &    0.162  &  0.173 & 0.691     \\
           Visual ChatGPT-4 \cite{wu2023visual-chatGPT} & 0.860  &  0.794       &   0.711   &  0.744 &  0.789  \\\rowcolor{Gray}
			\methodname  & \textbf{1.000}  &    \textbf{0.959}     &  \textbf{0.927}    & \textbf{0.955}  &  \textbf{0.962}    \\
		\end{tabular} 
	}
 \vspace{0.7mm}
    \small\caption{\footnotesize
	            Comparison of tool usage accuracy within multi-turn dialogue. Successful rate of thought ($\text{SR}_\text{t}$), action ($\text{SR}_\text{act}$), arguments ($\text{SR}_\text{args}$), execution ($\text{SR}$), and IoU are reported. 
	}
 % \vspace{-6.5mm}
	\label{tab:tool_usage_multiturn}
\end{table}

%% file: tables/A08_shape_prompt.tex
\begin{table*}[t]
    \centering
    \footnotesize
    \begin{tcolorbox} 
    \renewcommand\arraystretch{1.2} %
    \begin{tabular}{p{\textwidth}}
        You are an AI visual assistant, and you are seeing a single image. What you see are provided with a sentence, describing the body shape of the person in the image. Answer all questions as you are seeing the image. \\ \\

The sentence includes information about the person's gender, body mass, height, chest circumference, waist circumference, and hip circumference. Besides, it includes 15 linguistic shape attributes scale from 1 (strongly disagree) to 5 (strongly agree). \\

Design a conversation between you and a person asking about the body shape of the person. The answers should be in a tone that a visual AI assistant is seeing the image and answering the question. \\
Ask diverse questions and give corresponding answers. \\ \\

Include questions asking about the visual content of the image, including the person's overall body fit, shape, height, mass, etc. Only include questions that have definite answers: \\ 
(1) one can see the content in the image that the question asks about and can answer confidently; \\ 
(2) one can determine confidently from the image that it is not in the image.
Do not ask any questions that cannot be answered confidently. \\ \\

Provide detailed answers when answering complex questions. When the question is about the measurement, provide an explicit and concrete metric number in the answer. \\
In your answer, you should imitate as if you see the image and the measurements and linguistic attributes are estimated by you. The linguistic attribute score is only used to help you understand and don't mention it in your answer. \\ \\

Directly reply to me with a list, here are some examples: \\
1. How tall is the person in the image? [The person looks quite tall. He is about 1.85 m.] \\
2. Please help me estimate the body measurements of the man in the image. [The man is about 1.74 m and 60 kg. His chest circumference is about 0.9 m.] \\
3. What's the waist circumference of the person? [The chest circumference is about 0.95 m.]
    \end{tabular}
    \end{tcolorbox} 
    \caption{\footnotesize
        Prompt to request GPT-4 to generate instruction-following data about human body shape based on the textual description about human body.
    }
    \label{tab:shape_prompt}
    \end{table*}

%% file: tables/R05_tool_graph_construction.tex
\begin{table*}
\centering
    \footnotesize
    \begin{tcolorbox} 
    \renewcommand\arraystretch{1.2} %
    \begin{tabular}{p{\textwidth}}
    
    Given an image whose image path is "example.jpg". Image caption: "\{caption\}". The image caption includes detail image description and each object paired with the bounding box [x1, y1, x2, y2]. For the bounding box, (x1, y1) refers to the top left, and (x2, y2) refers to the bottom right. x1 less than x2, and y1 less than y2. \\ \\

Below are 26 visual tools. Each tool is defined as "tool name: usage scenario, and arguments to tool".\\ \\

Please generate 10 instructions that will need multiple tools to finish. 
The generated instructions should follow the format of "instruction content, [[tool name1, arguments to tool1], [tool name2, arguments to tool2], ...]".  Each instruction must relate to the caption and can be solved by the tool. 
You can not revise the "tool name", or add any other fake tools that is not defined. You must keep the correct "arguments to tool".\\ \\

Tools:
\{tool description\} \\ \\

Note that you should use 1-5 tools in each instruction and your generated visual instructions should be highly related to the image caption. Directly reply to me with the list, here are some examples:\\
\{examples\} \\ \\

Diversify the instructions to cover a wide range of possible user queries.  Feel free to adapt and rephrase the examples provided to generate diverse, complex, and deceptive instructions as much as possible. \\
For example, you can also change the subject position or the person and pose description positions. Don't use too much imperative sentence, you should also use interrogative sentence.

    \end{tabular}
    \end{tcolorbox} 
    \caption{\footnotesize
        Prompt to request GPT-4 to construct tool graph based on the image description, tool description, and tool usage examples summarized from tool paper.
    }
    \label{tab:tool_graph_construction}
    \end{table*}

%% file: main.bbl
\begin{thebibliography}{85}
\providecommand{\natexlab}[1]{#1}
\providecommand{\url}[1]{\texttt{#1}}
\expandafter\ifx\csname urlstyle\endcsname\relax
  \providecommand{\doi}[1]{doi: #1}\else
  \providecommand{\doi}{doi: \begingroup \urlstyle{rm}\Url}\fi

\bibitem[Aberman et~al.(2019)Aberman, Wu, Lischinski, Chen, and Cohen-Or]{aberman2019learning}
Kfir Aberman, Rundi Wu, Dani Lischinski, Baoquan Chen, and Daniel Cohen-Or.
\newblock Learning character-agnostic motion for motion retargeting in 2d.
\newblock \emph{arXiv preprint arXiv:1905.01680}, 2019.

\bibitem[Black et~al.(2023)Black, Patel, Tesch, and Yang]{bedlam}
Michael~J Black, Priyanka Patel, Joachim Tesch, and Jinlong Yang.
\newblock Bedlam: A synthetic dataset of bodies exhibiting detailed lifelike animated motion.
\newblock In \emph{Proceedings of the IEEE/CVF Conference on Computer Vision and Pattern Recognition}, pages 8726--8737, 2023.

\bibitem[Blattmann et~al.(2023)Blattmann, Dockhorn, Kulal, Mendelevitch, Kilian, Lorenz, Levi, English, Voleti, Letts, et~al.]{blattmann2023stable}
Andreas Blattmann, Tim Dockhorn, Sumith Kulal, Daniel Mendelevitch, Maciej Kilian, Dominik Lorenz, Yam Levi, Zion English, Vikram Voleti, Adam Letts, et~al.
\newblock Stable video diffusion: Scaling latent video diffusion models to large datasets.
\newblock \emph{arXiv preprint arXiv:2311.15127}, 2023.

\bibitem[Bogo et~al.(2016)Bogo, Kanazawa, Lassner, Gehler, Romero, and Black]{smplify}
Federica Bogo, Angjoo Kanazawa, Christoph Lassner, Peter Gehler, Javier Romero, and Michael~J. Black.
\newblock Keep it {SMPL}: {A}utomatic estimation of {3D} human pose and shape from a single image.
\newblock In \emph{ECCV}, 2016.

\bibitem[Cao et~al.(2023)Cao, Cao, Han, Shan, and Wong]{cao2023dreamavatar}
Yukang Cao, Yan-Pei Cao, Kai Han, Ying Shan, and Kwan-Yee~K Wong.
\newblock Dreamavatar: Text-and-shape guided 3d human avatar generation via diffusion models.
\newblock \emph{arXiv preprint arXiv:2304.00916}, 2023.

\bibitem[Chase and Contributors(2022)]{langchain}
Harrison Chase and LangChain Contributors.
\newblock Langchain, 2022.

\bibitem[Chiang et~al.(2023)Chiang, Li, Lin, Sheng, Wu, Zhang, Zheng, Zhuang, Zhuang, Gonzalez, Stoica, and Xing]{vicuna}
Wei-Lin Chiang, Zhuohan Li, Zi Lin, Ying Sheng, Zhanghao Wu, Hao Zhang, Lianmin Zheng, Siyuan Zhuang, Yonghao Zhuang, Joseph~E. Gonzalez, Ion Stoica, and Eric~P. Xing.
\newblock Vicuna: An open-source chatbot impressing gpt-4 with 90\%* chatgpt quality, 2023.

\bibitem[Choutas et~al.(2022)Choutas, M{\"u}ller, Huang, Tang, Tzionas, and Black]{shapy}
Vasileios Choutas, Lea M{\"u}ller, Chun-Hao~P Huang, Siyu Tang, Dimitrios Tzionas, and Michael~J Black.
\newblock Accurate 3d body shape regression using metric and semantic attributes.
\newblock In \emph{Proceedings of the IEEE/CVF Conference on Computer Vision and Pattern Recognition}, pages 2718--2728, 2022.

\bibitem[Dan{\v{e}}{\v{c}}ek et~al.(2022)Dan{\v{e}}{\v{c}}ek, Black, and Bolkart]{danvevcek2022emoca}
Radek Dan{\v{e}}{\v{c}}ek, Michael~J Black, and Timo Bolkart.
\newblock Emoca: Emotion driven monocular face capture and animation.
\newblock In \emph{Proceedings of the IEEE/CVF Conference on Computer Vision and Pattern Recognition}, pages 20311--20322, 2022.

\bibitem[Delmas et~al.(2022)Delmas, Weinzaepfel, Lucas, Moreno-Noguer, and Rogez]{posescript}
Ginger Delmas, Philippe Weinzaepfel, Thomas Lucas, Francesc Moreno-Noguer, and Gr{\'e}gory Rogez.
\newblock Posescript: 3d human poses from natural language.
\newblock In \emph{ECCV}, 2022.

\bibitem[Delmas et~al.(2023)Delmas, Weinzaepfel, Moreno-Noguer, and Rogez]{delmas2023posefix}
Ginger Delmas, Philippe Weinzaepfel, Francesc Moreno-Noguer, and Gr{\'e}gory Rogez.
\newblock Posefix: Correcting 3d human poses with natural language.
\newblock In \emph{Proceedings of the IEEE/CVF International Conference on Computer Vision}, pages 15018--15028, 2023.

\bibitem[Deng et~al.(2019)Deng, Yang, Xu, Chen, Jia, and Tong]{deng2019accurate}
Yu Deng, Jiaolong Yang, Sicheng Xu, Dong Chen, Yunde Jia, and Xin Tong.
\newblock Accurate 3d face reconstruction with weakly-supervised learning: From single image to image set.
\newblock In \emph{IEEE Computer Vision and Pattern Recognition Workshops}, 2019.

\bibitem[Du et~al.(2024)Du, Wei, and Zhang]{AnyTool}
Yu Du, Fangyun Wei, and Hongyang Zhang.
\newblock Anytool: Self-reflective, hierarchical agents for large-scale api calls.
\newblock \emph{arXiv preprint arXiv:2402.04253}, 2024.

\bibitem[Feng et~al.(2021{\natexlab{a}})Feng, Choutas, Bolkart, Tzionas, and Black]{pixie}
Yao Feng, Vasileios Choutas, Timo Bolkart, Dimitrios Tzionas, and Michael~J. Black.
\newblock Collaborative regression of expressive bodies using moderation.
\newblock In \emph{{3DV}}, 2021{\natexlab{a}}.

\bibitem[Feng et~al.(2021{\natexlab{b}})Feng, Feng, Black, and Bolkart]{DECA:Siggraph2021}
Yao Feng, Haiwen Feng, Michael~J Black, and Timo Bolkart.
\newblock Learning an animatable detailed 3d face model from in-the-wild images.
\newblock \emph{ACM Transactions on Graphics}, 40\penalty0 (4):\penalty0 1--13, 2021{\natexlab{b}}.

\bibitem[Feng et~al.(2024)Feng, Lin, Dwivedi, Sun, Patel, and Black]{feng2024chatpose}
Yao Feng, Jing Lin, Sai~Kumar Dwivedi, Yu Sun, Priyanka Patel, and Michael~J. Black.
\newblock {ChatPose}: Chatting about 3d human pose.
\newblock In \emph{CVPR}, 2024.

\bibitem[Gao et~al.(2023)Gao, Xiong, Gao, Jia, Pan, Bi, Dai, Sun, and Wang]{gao2023retrieval}
Yunfan Gao, Yun Xiong, Xinyu Gao, Kangxiang Jia, Jinliu Pan, Yuxi Bi, Yi Dai, Jiawei Sun, and Haofen Wang.
\newblock Retrieval-augmented generation for large language models: A survey.
\newblock \emph{arXiv preprint arXiv:2312.10997}, 2023.

\bibitem[Goel et~al.(2023)Goel, Pavlakos, Rajasegaran, Kanazawa, and Malik]{hmr2}
Shubham Goel, Georgios Pavlakos, Jathushan Rajasegaran, Angjoo Kanazawa, and Jitendra Malik.
\newblock Humans in 4{D}: Reconstructing and tracking humans with transformers.
\newblock In \emph{ICCV}, 2023.

\bibitem[Han and Joo(2023)]{han2023chorus}
Sookwan Han and Hanbyul Joo.
\newblock Chorus: Learning canonicalized 3d human-object spatial relations from unbounded synthesized images.
\newblock In \emph{Proceedings of the IEEE/CVF International Conference on Computer Vision}, pages 15835--15846, 2023.

\bibitem[Hong et~al.(2022)Hong, Zhang, Pan, Cai, Yang, and Liu]{hong2022avatarclip}
Fangzhou Hong, Mingyuan Zhang, Liang Pan, Zhongang Cai, Lei Yang, and Ziwei Liu.
\newblock Avatarclip: Zero-shot text-driven generation and animation of 3d avatars.
\newblock \emph{arXiv preprint arXiv:2205.08535}, 2022.

\bibitem[Hsieh et~al.(2023)Hsieh, Chen, Li, Fujii, Ratner, Lee, Krishna, and Pfister]{hsieh2023tool}
Cheng-Yu Hsieh, Si-An Chen, Chun-Liang Li, Yasuhisa Fujii, Alexander Ratner, Chen-Yu Lee, Ranjay Krishna, and Tomas Pfister.
\newblock Tool documentation enables zero-shot tool-usage with large language models.
\newblock \emph{arXiv preprint arXiv:2308.00675}, 2023.

\bibitem[Hu et~al.(2021)Hu, Shen, Wallis, Allen-Zhu, Li, Wang, Wang, and Chen]{hu2021lora}
Edward~J Hu, Yelong Shen, Phillip Wallis, Zeyuan Allen-Zhu, Yuanzhi Li, Shean Wang, Lu Wang, and Weizhu Chen.
\newblock Lora: Low-rank adaptation of large language models.
\newblock \emph{arXiv:2106.09685}, 2021.

\bibitem[Jiang et~al.(2023)Jiang, Won, Ye, and Liu]{jiang2023drop}
Yifeng Jiang, Jungdam Won, Yuting Ye, and C~Karen Liu.
\newblock Drop: Dynamics responses from human motion prior and projective dynamics.
\newblock \emph{SIGGRAPH Asia}, 2023.

\bibitem[Joo et~al.(2020)Joo, Neverova, and Vedaldi]{eft}
Hanbyul Joo, Natalia Neverova, and Andrea Vedaldi.
\newblock Exemplar fine-tuning for {3D} human pose fitting towards in-the-wild {3D} human pose estimation.
\newblock In \emph{{3DV}}, 2020.

\bibitem[Kanazawa et~al.(2018)Kanazawa, Black, Jacobs, and Malik]{hmr}
Angjoo Kanazawa, Michael~J. Black, David~W. Jacobs, and Jitendra Malik.
\newblock End-to-end recovery of human shape and pose.
\newblock In \emph{CVPR}, 2018.

\bibitem[Kirillov et~al.(2023)Kirillov, Mintun, Ravi, Mao, Rolland, Gustafson, Xiao, Whitehead, Berg, Lo, et~al.]{kirillov2023segment}
Alexander Kirillov, Eric Mintun, Nikhila Ravi, Hanzi Mao, Chloe Rolland, Laura Gustafson, Tete Xiao, Spencer Whitehead, Alexander~C Berg, Wan-Yen Lo, et~al.
\newblock Segment anything.
\newblock In \emph{Proceedings of the IEEE/CVF International Conference on Computer Vision}, pages 4015--4026, 2023.

\bibitem[Kolotouros et~al.(2019)Kolotouros, Pavlakos, Black, and Daniilidis]{spin}
Nikos Kolotouros, Georgios Pavlakos, Michael~J. Black, and Kostas Daniilidis.
\newblock Learning to reconstruct {3D} human pose and shape via model-fitting in the loop.
\newblock In \emph{ICCV}, pages 2252--2261, 2019.

\bibitem[Kong et~al.(2023{\natexlab{a}})Kong, Ruan, Chen, Zhang, Bao, Shi, Du, Hu, Mao, Li, et~al.]{TPTU-v2}
Yilun Kong, Jingqing Ruan, Yihong Chen, Bin Zhang, Tianpeng Bao, Shiwei Shi, Guoqing Du, Xiaoru Hu, Hangyu Mao, Ziyue Li, et~al.
\newblock Tptu-v2: Boosting task planning and tool usage of large language model-based agents in real-world systems.
\newblock \emph{arXiv preprint arXiv:2311.11315}, 2023{\natexlab{a}}.

\bibitem[Kong et~al.(2023{\natexlab{b}})Kong, Ruan, Chen, Zhang, Bao, Shi, Du, Hu, Mao, Li, et~al.]{kong2023tptu}
Yilun Kong, Jingqing Ruan, Yihong Chen, Bin Zhang, Tianpeng Bao, Shiwei Shi, Guoqing Du, Xiaoru Hu, Hangyu Mao, Ziyue Li, et~al.
\newblock Tptu-v2: Boosting task planning and tool usage of large language model-based agents in real-world systems.
\newblock \emph{arXiv preprint arXiv:2311.11315}, 2023{\natexlab{b}}.

\bibitem[Lewis et~al.(2020{\natexlab{a}})Lewis, Perez, Piktus, Petroni, Karpukhin, Goyal, K{\"u}ttler, Lewis, Yih, Rockt{\"a}schel, et~al.]{RAG}
Patrick Lewis, Ethan Perez, Aleksandra Piktus, Fabio Petroni, Vladimir Karpukhin, Naman Goyal, Heinrich K{\"u}ttler, Mike Lewis, Wen-tau Yih, Tim Rockt{\"a}schel, et~al.
\newblock Retrieval-augmented generation for knowledge-intensive nlp tasks.
\newblock \emph{Advances in neural information processing systems}, 33:\penalty0 9459--9474, 2020{\natexlab{a}}.

\bibitem[Lewis et~al.(2020{\natexlab{b}})Lewis, Perez, Piktus, Petroni, Karpukhin, Goyal, K{\"u}ttler, Lewis, Yih, Rockt{\"a}schel, et~al.]{lewis2020retrieval}
Patrick Lewis, Ethan Perez, Aleksandra Piktus, Fabio Petroni, Vladimir Karpukhin, Naman Goyal, Heinrich K{\"u}ttler, Mike Lewis, Wen-tau Yih, Tim Rockt{\"a}schel, et~al.
\newblock Retrieval-augmented generation for knowledge-intensive nlp tasks.
\newblock \emph{Advances in Neural Information Processing Systems}, 33:\penalty0 9459--9474, 2020{\natexlab{b}}.

\bibitem[Li et~al.(2021{\natexlab{a}})Li, Xu, Chen, Bian, Yang, and Lu]{hybrik}
Jiefeng Li, Chao Xu, Zhicun Chen, Siyuan Bian, Lixin Yang, and Cewu Lu.
\newblock {HybrIK}: {A} hybrid analytical-neural inverse kinematics solution for {3D} human pose and shape estimation.
\newblock In \emph{CVPR}, 2021{\natexlab{a}}.

\bibitem[Li et~al.(2023)Li, Wu, and Liu]{li2023object}
Jiaman Li, Jiajun Wu, and C~Karen Liu.
\newblock Object motion guided human motion synthesis.
\newblock \emph{ACM Transactions on Graphics (TOG)}, 42\penalty0 (6):\penalty0 1--11, 2023.

\bibitem[Li et~al.(2021{\natexlab{b}})Li, Aberman, Hanocka, Liu, Sorkine-Hornung, and Chen]{li2021learning}
Peizhuo Li, Kfir Aberman, Rana Hanocka, Libin Liu, Olga Sorkine-Hornung, and Baoquan Chen.
\newblock Learning skeletal articulations with neural blend shapes.
\newblock \emph{ACM Transactions on Graphics (TOG)}, 40\penalty0 (4):\penalty0 1, 2021{\natexlab{b}}.

\bibitem[Li et~al.(2017)Li, Bolkart, Black, Li, and Romero]{FLAME:SiggraphAsia2017}
Tianye Li, Timo Bolkart, Michael.~J. Black, Hao Li, and Javier Romero.
\newblock Learning a model of facial shape and expression from {4D} scans.
\newblock \emph{ACM Transactions on Graphics, (Proc. SIGGRAPH Asia)}, 36\penalty0 (6):\penalty0 194:1--194:17, 2017.

\bibitem[Li et~al.(2022)Li, Liu, Zhang, Xu, and Yan]{cliff}
Zhihao Li, Jianzhuang Liu, Zhensong Zhang, Songcen Xu, and Youliang Yan.
\newblock {CLIFF}: {C}arrying location information in full frames into human pose and shape estimation.
\newblock In \emph{ECCV}, 2022.

\bibitem[Lin et~al.(2023)Lin, Zeng, Wang, Zhang, and Li]{lin2023osx}
Jing Lin, Ailing Zeng, Haoqian Wang, Lei Zhang, and Yu Li.
\newblock One-stage 3d whole-body mesh recovery with component aware transformer.
\newblock \emph{CVPR}, 2023.

\bibitem[Lin et~al.(2021)Lin, Wang, and Liu]{hamer}
Kevin Lin, Lijuan Wang, and Zicheng Liu.
\newblock End-to-end human pose and mesh reconstruction with transformers.
\newblock In \emph{Proceedings of the IEEE/CVF conference on computer vision and pattern recognition}, pages 1954--1963, 2021.

\bibitem[Lin et~al.(2014)Lin, Maire, Belongie, Hays, Perona, Ramanan, Doll{\'a}r, and Zitnick]{coco}
Tsung-Yi Lin, Michael Maire, Serge~J. Belongie, James Hays, Pietro Perona, Deva Ramanan, Piotr Doll{\'a}r, and C.~Lawrence Zitnick.
\newblock Microsoft coco: Common objects in context.
\newblock In \emph{ECCV}, 2014.

\bibitem[Liu et~al.(2023{\natexlab{a}})Liu, Li, Wu, and Lee]{liu2023visual}
Haotian Liu, Chunyuan Li, Qingyang Wu, and Yong~Jae Lee.
\newblock Visual instruction tuning.
\newblock \emph{arXiv preprint arXiv:2304.08485}, 2023{\natexlab{a}}.

\bibitem[Liu et~al.(2023{\natexlab{b}})Liu, Li, Wu, and Lee]{llava}
Haotian Liu, Chunyuan Li, Qingyang Wu, and Yong~Jae Lee.
\newblock Visual instruction tuning.
\newblock In \emph{NeurIPS}, 2023{\natexlab{b}}.

\bibitem[Liu et~al.(2023{\natexlab{c}})Liu, Cheng, Liu, Zhang, Li, Ren, Zou, Yang, Su, Zhu, Zhang, Gao, and Li]{liu2023llavaplus}
Shilong Liu, Hao Cheng, Haotian Liu, Hao Zhang, Feng Li, Tianhe Ren, Xueyan Zou, Jianwei Yang, Hang Su, Jun Zhu, Lei Zhang, Jianfeng Gao, and Chunyuan Li.
\newblock Llava-plus: Learning to use tools for creating multimodal agents, 2023{\natexlab{c}}.

\bibitem[Loper et~al.(2015)Loper, Mahmood, Romero, Pons-Moll, and Black]{smpl}
Matthew Loper, Naureen Mahmood, Javier Romero, Gerard Pons-Moll, and Michael~J. Black.
\newblock {SMPL}: {A} skinned multi-person linear model.
\newblock In \emph{ACM TOG}, 2015.

\bibitem[Loshchilov and Hutter(2017)]{adamw}
Ilya Loshchilov and Frank Hutter.
\newblock Decoupled weight decay regularization.
\newblock \emph{arXiv preprint arXiv:1711.05101}, 2017.

\bibitem[Lucas* et~al.(2022)Lucas*, Baradel*, Weinzaepfel, and Rogez]{posegpt}
Thomas Lucas*, Fabien Baradel*, Philippe Weinzaepfel, and Gr\'egory Rogez.
\newblock Posegpt: Quantization-based 3d human motion generation and forecasting.
\newblock In \emph{European Conference on Computer Vision ({ECCV})}, 2022.

\bibitem[Matl(2019)]{pyrender}
Matthew Matl.
\newblock Pyrender.
\newblock \url{https://github.com/mmatl/pyrender}, 2019.

\bibitem[Muller et~al.(2021)Muller, Osman, Tang, Huang, and Black]{muller2021self}
Lea Muller, Ahmed~AA Osman, Siyu Tang, Chun-Hao~P Huang, and Michael~J Black.
\newblock On self-contact and human pose.
\newblock In \emph{Proceedings of the IEEE/CVF Conference on Computer Vision and Pattern Recognition}, pages 9990--9999, 2021.

\bibitem[OpenAI(2023)]{gpt4}
OpenAI.
\newblock {GPT-4} technical report.
\newblock 2023.

\bibitem[Pavlakos et~al.(2019)Pavlakos, Choutas, Ghorbani, Bolkart, Osman, Tzionas, and Black]{smplx}
Georgios Pavlakos, Vasileios Choutas, Nima Ghorbani, Timo Bolkart, Ahmed A.~A. Osman, Dimitrios Tzionas, and Michael~J. Black.
\newblock Expressive body capture: {3D} hands, face, and body from a single image.
\newblock In \emph{CVPR}, 2019.

\bibitem[Paysan et~al.(2009)Paysan, Knothe, Amberg, Romdhani, and Vetter]{bfm09}
P. Paysan, R. Knothe, B. Amberg, S. Romdhani, and T. Vetter.
\newblock A {3D} face model for pose and illumination invariant face recognition.
\newblock Genova, Italy, 2009. IEEE.

\bibitem[Petrovich et~al.(2023)Petrovich, Black, and Varol]{petrovich23tmr}
Mathis Petrovich, Michael~J. Black, and G{\"u}l Varol.
\newblock {TMR}: Text-to-motion retrieval using contrastive {3D} human motion synthesis.
\newblock In \emph{International Conference on Computer Vision ({ICCV})}, 2023.

\bibitem[Radford et~al.(2021)Radford, Kim, Hallacy, Ramesh, Goh, Agarwal, Sastry, Askell, Mishkin, Clark, et~al.]{clip}
Alec Radford, Jong~Wook Kim, Chris Hallacy, Aditya Ramesh, Gabriel Goh, Sandhini Agarwal, Girish Sastry, Amanda Askell, Pamela Mishkin, Jack Clark, et~al.
\newblock Learning transferable visual models from natural language supervision.
\newblock In \emph{International conference on machine learning}, pages 8748--8763. PMLR, 2021.

\bibitem[Rasley et~al.(2020)Rasley, Rajbhandari, Ruwase, and He]{rasley2020deepspeed}
Jeff Rasley, Samyam Rajbhandari, Olatunji Ruwase, and Yuxiong He.
\newblock Deepspeed: System optimizations enable training deep learning models with over 100 billion parameters.
\newblock \emph{arXiv preprint arXiv:2002.11681}, 2020.

\bibitem[Rasley et~al.(2023)Rasley, Rajbhandari, Ruwase, and He]{wu2023visual-chatGPT}
Jeff Rasley, Samyam Rajbhandari, Olatunji Ruwase, and Yuxiong He.
\newblock Visual chatgpt: Talking, drawing and editing with visual foundation models.
\newblock \emph{arXiv preprint arXiv:2303.04671}, 2023.

\bibitem[Rombach et~al.(2022)Rombach, Blattmann, Lorenz, Esser, and Ommer]{stable_diffusion}
Robin Rombach, Andreas Blattmann, Dominik Lorenz, Patrick Esser, and Bj{\"o}rn Ommer.
\newblock High-resolution image synthesis with latent diffusion models.
\newblock In \emph{Proceedings of the IEEE/CVF conference on computer vision and pattern recognition}, pages 10684--10695, 2022.

\bibitem[Romero et~al.(2017)Romero, Tzionas, and Black]{MANO:SIGGRAPHASIA:2017}
Javier Romero, Dimitrios Tzionas, and Michael~J. Black.
\newblock Embodied hands: Modeling and capturing hands and bodies together.
\newblock \emph{ACM Transactions on Graphics, (Proc. SIGGRAPH Asia)}, 36\penalty0 (6), 2017.

\bibitem[Rong et~al.(2021)Rong, Shiratori, and Joo]{rong2021frankmocap}
Yu Rong, Takaaki Shiratori, and Hanbyul Joo.
\newblock Frankmocap: A monocular 3d whole-body pose estimation system via regression and integration.
\newblock In \emph{IEEE International Conference on Computer Vision Workshops}, 2021.

\bibitem[Ruan et~al.(2023)Ruan, Chen, Zhang, Xu, Bao, Mao, Li, Zeng, Zhao, et~al.]{ruan2023tptu}
Jingqing Ruan, Yihong Chen, Bin Zhang, Zhiwei Xu, Tianpeng Bao, Hangyu Mao, Ziyue Li, Xingyu Zeng, Rui Zhao, et~al.
\newblock Tptu: Task planning and tool usage of large language model-based ai agents.
\newblock In \emph{NeurIPS 2023 Foundation Models for Decision Making Workshop}, 2023.

\bibitem[Shen et~al.(2023{\natexlab{a}})Shen, Song, Tan, Li, Lu, and Zhuang]{shen2023hugginggpt}
Yongliang Shen, Kaitao Song, Xu Tan, Dongsheng Li, Weiming Lu, and Yueting Zhuang.
\newblock Hugginggpt: Solving ai tasks with chatgpt and its friends in huggingface.
\newblock \emph{arXiv preprint arXiv:2303.17580}, 2023{\natexlab{a}}.

\bibitem[Shen et~al.(2023{\natexlab{b}})Shen, Song, Tan, Zhang, Ren, Yuan, Lu, Li, and Zhuang]{shen2023taskbench}
Yongliang Shen, Kaitao Song, Xu Tan, Wenqi Zhang, Kan Ren, Siyu Yuan, Weiming Lu, Dongsheng Li, and Yueting Zhuang.
\newblock Taskbench: Benchmarking large language models for task automation.
\newblock \emph{arXiv preprint arXiv:2311.18760}, 2023{\natexlab{b}}.

\bibitem[Shin et~al.(2024)Shin, Kim, Halilaj, and Black]{shin2023wham}
Soyong Shin, Juyong Kim, Eni Halilaj, and Michael~J. Black.
\newblock Wham: Reconstructing world-grounded humans with accurate 3d motion.
\newblock In \emph{CVPR}, 2024.

\bibitem[Su et~al.(2022)Su, Shi, Kasai, Wang, Hu, Ostendorf, Yih, Smith, Zettlemoyer, and Yu]{su2022one}
Hongjin Su, Weijia Shi, Jungo Kasai, Yizhong Wang, Yushi Hu, Mari Ostendorf, Wen-tau Yih, Noah~A Smith, Luke Zettlemoyer, and Tao Yu.
\newblock One embedder, any task: Instruction-finetuned text embeddings.
\newblock \emph{arXiv preprint arXiv:2212.09741}, 2022.

\bibitem[Sur\'is et~al.(2023)Sur\'is, Menon, and Vondrick]{surismenon2023vipergpt}
D\'idac Sur\'is, Sachit Menon, and Carl Vondrick.
\newblock Vipergpt: Visual inference via python execution for reasoning.
\newblock \emph{Proceedings of IEEE International Conference on Computer Vision (ICCV)}, 2023.

\bibitem[Sweller et~al.(2011)Sweller, Ayres, and Kalyuga]{Sweller2011CognitiveLT}
John Sweller, Paul Ayres, and Slava Kalyuga.
\newblock \emph{Cognitive Load Theory}.
\newblock Springer, New York, NY, 2011.

\bibitem[Tevet et~al.(2022)Tevet, Gordon, Hertz, Bermano, and Cohen-Or]{tevet2022motionclip}
Guy Tevet, Brian Gordon, Amir Hertz, Amit~H Bermano, and Daniel Cohen-Or.
\newblock Motionclip: Exposing human motion generation to clip space.
\newblock In \emph{Computer Vision--ECCV 2022: 17th European Conference, Tel Aviv, Israel, October 23--27, 2022, Proceedings, Part XXII}, pages 358--374. Springer, 2022.

\bibitem[Tevet et~al.(2023)Tevet, Raab, Gordon, Shafir, Cohen-or, and Bermano]{tevet2023human}
Guy Tevet, Sigal Raab, Brian Gordon, Yoni Shafir, Daniel Cohen-or, and Amit~Haim Bermano.
\newblock Human motion diffusion model.
\newblock In \emph{The Eleventh International Conference on Learning Representations}, 2023.

\bibitem[Tewari et~al.(2017)Tewari, Zollhofer, Kim, Garrido, Bernard, Perez, and Theobalt]{tewari2017mofa}
Ayush Tewari, Michael Zollhofer, Hyeongwoo Kim, Pablo Garrido, Florian Bernard, Patrick Perez, and Christian Theobalt.
\newblock Mofa: Model-based deep convolutional face autoencoder for unsupervised monocular reconstruction.
\newblock In \emph{Proceedings of the IEEE international conference on computer vision workshops}, pages 1274--1283, 2017.

\bibitem[Touvron et~al.(2023)Touvron, Martin, Stone, Albert, Almahairi, Babaei, Bashlykov, Batra, Bhargava, Bhosale, et~al.]{touvron2023llama}
Hugo Touvron, Louis Martin, Kevin Stone, Peter Albert, Amjad Almahairi, Yasmine Babaei, Nikolay Bashlykov, Soumya Batra, Prajjwal Bhargava, Shruti Bhosale, et~al.
\newblock Llama 2: Open foundation and fine-tuned chat models.
\newblock \emph{arXiv preprint arXiv:2307.09288}, 2023.

\bibitem[Tripathi et~al.(2023{\natexlab{a}})Tripathi, Chatterjee, Passy, Yi, Tzionas, and Black]{deco}
Shashank Tripathi, Agniv Chatterjee, Jean-Claude Passy, Hongwei Yi, Dimitrios Tzionas, and Michael~J Black.
\newblock Deco: Dense estimation of 3d human-scene contact in the wild.
\newblock In \emph{Proceedings of the IEEE/CVF International Conference on Computer Vision}, pages 8001--8013, 2023{\natexlab{a}}.

\bibitem[Tripathi et~al.(2023{\natexlab{b}})Tripathi, M{\"u}ller, Huang, Taheri, Black, and Tzionas]{moyo}
Shashank Tripathi, Lea M{\"u}ller, Chun-Hao~P Huang, Omid Taheri, Michael~J Black, and Dimitrios Tzionas.
\newblock 3d human pose estimation via intuitive physics.
\newblock In \emph{Proceedings of the IEEE/CVF Conference on Computer Vision and Pattern Recognition}, pages 4713--4725, 2023{\natexlab{b}}.

\bibitem[von Marcard et~al.(2018)von Marcard, Henschel, Black, Rosenhahn, and Pons-Moll]{threedpw}
Timo von Marcard, Roberto Henschel, Michael~J. Black, Bodo Rosenhahn, and Gerard Pons-Moll.
\newblock Recovering accurate {3D} human pose in the wild using {IMUs} and a moving camera.
\newblock In \emph{ECCV}, 2018.

\bibitem[Wang et~al.(2024)Wang, Luo, Chen, Mai, Guo, Dong, Xuan, Li, Ma, and Gao]{wang2024mllmtool}
Chenyu Wang, Weixin Luo, Qianyu Chen, Haonan Mai, Jindi Guo, Sixun Dong, Xiaohua~(Michael) Xuan, Zhengxin Li, Lin Ma, and Shenghua Gao.
\newblock Mllm-tool: A multimodal large language model for tool agent learning.
\newblock \emph{arXiv preprint arXiv:2401.10727}, 2024.

\bibitem[Wu et~al.(2024)Wu, Liu, Luan, and Wang]{wu2024toolplanner}
Qinzhuo Wu, Wei Liu, Jian Luan, and Bin Wang.
\newblock Toolplanner: A tool augmented llm for multi granularity instructions with path planning and feedback.
\newblock \emph{arXiv preprint arXiv:2409.14826}, 2024.

\bibitem[Xu et~al.(2020)Xu, Bazavan, Zanfir, Freeman, Sukthankar, and Sminchisescu]{xu2020ghum}
Hongyi Xu, Eduard~Gabriel Bazavan, Andrei Zanfir, William~T. Freeman, Rahul Sukthankar, and Cristian Sminchisescu.
\newblock {GHUM} {\&} {GHUML}: Generative {3D} human shape and articulated pose models.
\newblock In \emph{CVPR}, 2020.

\bibitem[Yang et~al.(2023{\natexlab{a}})Yang, Song, Li, Zhao, Ge, Li, and Shan]{gpt4tools}
Rui Yang, Lin Song, Yanwei Li, Sijie Zhao, Yixiao Ge, Xiu Li, and Ying Shan.
\newblock {GPT4Tools}: Teaching llm to use tools via self-instruction.
\newblock \emph{arXiv preprint arXiv:2305.18752}, 2023{\natexlab{a}}.

\bibitem[Yang et~al.(2023{\natexlab{b}})Yang, Song, Li, Zhao, Ge, Li, and Shan]{yang2023gpt4tools}
Rui Yang, Lin Song, Yanwei Li, Sijie Zhao, Yixiao Ge, Xiu Li, and Ying Shan.
\newblock Gpt4tools: Teaching large language model to use tools via self-instruction, 2023{\natexlab{b}}.

\bibitem[Yang et~al.(2023{\natexlab{c}})Yang, Li, Wang, Lin, Azarnasab, Ahmed, Liu, Liu, Zeng, and Wang]{yang2023mm}
Zhengyuan Yang, Linjie Li, Jianfeng Wang, Kevin Lin, Ehsan Azarnasab, Faisal Ahmed, Zicheng Liu, Ce Liu, Michael Zeng, and Lijuan Wang.
\newblock Mm-react: Prompting chatgpt for multimodal reasoning and action.
\newblock \emph{arXiv preprint arXiv:2303.11381}, 2023{\natexlab{c}}.

\bibitem[Yang et~al.(2023{\natexlab{d}})Yang, Liu, Han, Chen, Huang, Fu, and Yu]{yang2023appagent}
Zhao Yang, Jiaxuan Liu, Yucheng Han, Xin Chen, Zebiao Huang, Bin Fu, and Gang Yu.
\newblock {Appagent}: Multimodal agents as smartphone users.
\newblock \emph{arXiv preprint arXiv:2312.13771}, 2023{\natexlab{d}}.

\bibitem[Ye et~al.(2023)Ye, Lauer, Zhou, Mathis, and Mathis]{ye2023amadeusgpt}
Shaokai Ye, Jessy Lauer, Mu Zhou, Alexander Mathis, and Mackenzie Mathis.
\newblock Amadeusgpt: a natural language interface for interactive animal behavioral analysis.
\newblock \emph{Advances in neural information processing systems}, 36:\penalty0 6297--6329, 2023.

\bibitem[Ye et~al.(2024)Ye, Lauer, Zhou, Mathis, and Mathis]{ye2024amadeusgpt}
Shaokai Ye, Jessy Lauer, Mu Zhou, Alexander Mathis, and Mackenzie Mathis.
\newblock Amadeusgpt: a natural language interface for interactive animal behavioral analysis.
\newblock \emph{Advances in neural information processing systems}, 36, 2024.

\bibitem[Yuan et~al.(2024)Yuan, Song, Chen, Tan, Shen, Kan, Li, and Yang]{yuan2024easytool}
Siyu Yuan, Kaitao Song, Jiangjie Chen, Xu Tan, Yongliang Shen, Ren Kan, Dongsheng Li, and Deqing Yang.
\newblock Easytool: Enhancing llm-based agents with concise tool instruction.
\newblock \emph{arXiv preprint arXiv:2401.06201}, 2024.

\bibitem[Zhang et~al.(2021)Zhang, Tian, Zhou, Ouyang, Liu, Wang, and Sun]{zhang2021pymaf}
Hongwen Zhang, Yating Tian, Xinchi Zhou, Wanli Ouyang, Yebin Liu, Limin Wang, and Zhenan Sun.
\newblock Pymaf: 3d human pose and shape regression with pyramidal mesh alignment feedback loop.
\newblock In \emph{Proceedings of the IEEE/CVF International Conference on Computer Vision}, pages 11446--11456, 2021.

\bibitem[Zhang et~al.(2023)Zhang, Qiu, Lin, Zhang, Shi, Yang, Shi, Yang, Xu, and Yu]{zhang2023dreamface}
Longwen Zhang, Qiwei Qiu, Hongyang Lin, Qixuan Zhang, Cheng Shi, Wei Yang, Ye Shi, Sibei Yang, Lan Xu, and Jingyi Yu.
\newblock Dreamface: Progressive generation of animatable 3d faces under text guidance, 2023.

\bibitem[Zhao et~al.(2023)Zhao, Chen, Wang, Jiao, Do, Qin, Ding, Guo, Li, Li, et~al.]{zhao2023retrieving}
Ruochen Zhao, Hailin Chen, Weishi Wang, Fangkai Jiao, Xuan~Long Do, Chengwei Qin, Bosheng Ding, Xiaobao Guo, Minzhi Li, Xingxuan Li, et~al.
\newblock Retrieving multimodal information for augmented generation: A survey.
\newblock \emph{arXiv preprint arXiv:2303.10868}, 2023.

\bibitem[Zhu et~al.(2024)Zhu, Chen, Dai, Xu, Cao, Yao, Zhu, and Zhu]{zhu2024champ}
Shenhao Zhu, Junming~Leo Chen, Zuozhuo Dai, Yinghui Xu, Xun Cao, Yao Yao, Hao Zhu, and Siyu Zhu.
\newblock Champ: Controllable and consistent human image animation with 3d parametric guidance, 2024.

\end{thebibliography}
